\documentclass[10pt,twocolumn,letterpaper]{article}

\usepackage{cvpr}              % To produce the CAMERA-READY version
\usepackage[dvipsnames]{xcolor}

\definecolor{cvprblue}{rgb}{0.21,0.49,0.74}
\usepackage[pagebackref,breaklinks,colorlinks,citecolor=cvprblue]{hyperref}
\usepackage{bm}
\def\paperID{8085} % *** Enter the Paper ID here
\def\confName{CVPR}
\def\confYear{2024}

\title{Shadow Generation for Composite Image Using Diffusion Model}

\author{$\textnormal{Qingyang Liu}^{1}$, $\textnormal{Junqi You}^{1}$, $\textnormal{Jianting Wang}^{1}$, $\textnormal{Xinhao Tao}^{1}$, $\textnormal{Bo Zhang}^{1}$, $\textnormal{Li Niu}^{1,2}$\thanks{Corresponding author.}\\
$^1$ Shanghai Jiao Tong University 
$^2$ \href{https://miguocomics.com/}{miguo.ai}
\\
\tt\small$^1$\{narumimaria,yjqsjtu2022,glory1299,taoxinhao,bo-zhang,ustcnewly\}@sjtu.edu.cn 
}

\begin{document}
\maketitle
\begin{abstract}
In the realm of image composition, generating realistic shadow for the inserted  foreground remains a formidable challenge. Previous works have developed image-to-image translation models which are trained on paired training data. However, they are struggling to generate shadows with accurate shapes and intensities, hindered by data scarcity and inherent task complexity.  In this paper, we resort to foundation model with rich prior knowledge of natural shadow images. Specifically, we first adapt ControlNet to our task and then propose intensity modulation modules to improve the shadow intensity.  Moreover, we extend the small-scale DESOBA dataset to DESOBAv2 using a novel data acquisition pipeline. Experimental results on both DESOBA and DESOBAv2 datasets as well as real composite images demonstrate the superior capability of our model for shadow generation task. The dataset, code, and model are released at \href{https://github.com/bcmi/Object-Shadow-Generation-Dataset-DESOBAv2}{https://github.com/bcmi/Object-Shadow-Generation-Dataset-DESOBAv2}.

\end{abstract}    
\section{Introduction}
\label{sec:intro}

Image composition~\cite{niu2021making} aims to merge the foreground of one image with another background image to produce a composite image, which has a wide range of applications like virtual reality, artistic creation, and E-commerce. Simply pasting the foreground onto the background often results in visual inconsistencies, including the incompatible illumination between foreground and background~\cite{DoveNet2020}, lack of foreground shadow/reflection~\cite{hong2021shadow,sheng2021ssn}, and so on. In this paper, we focus on the shadow issue, \emph{i.e.}, the inserted foreground does not have plausible shadow on the background, which could significantly degrade the realism and quality of composite image. 

\begin{figure}[t]
\centering
\includegraphics[width=1.0\linewidth]{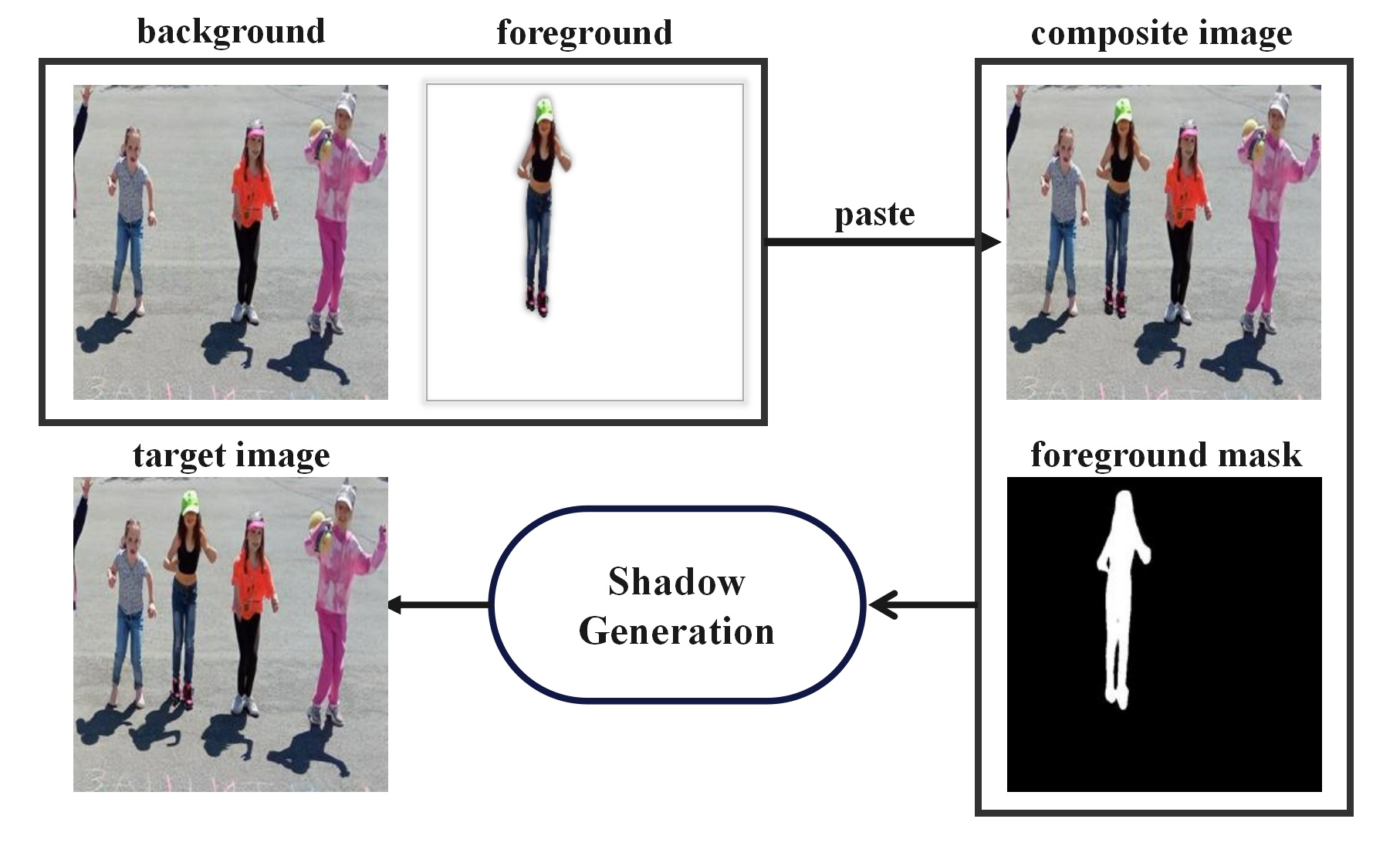}
\caption{A composite image can be obtained by pasting the foreground on the background. Shadow generation aims to generate plausible shadow for the inserted foreground in the composite image to produce a more realistic image.}
\label{fig:task_illustration}
\end{figure}

As illustrated in Figure \ref{fig:task_illustration}, shadow generation is a challenging task because the foreground shadow is determined by many complicated factors like the lighting information and the geometry of foreground/background. 
The existing shadow generation methods can be divided into rendering based methods \cite{sheng2021ssn,sheng2022controllable,sheng2023pixht} and non-rendering based methods \cite{zhang2019shadowgan,liu2020arshadowgan,hong2021shadow}. Rendering based methods usually impose restrict assumptions on the geometry and lighting, which could hardly be satisfied in real-world scenarios. Besides, \cite{sheng2022controllable,sheng2023pixht} require users to specify the lighting information, which hinders its direct application in our task. 
Non-rendering based methods usually train an image-to-image translation network, based on pairs of composite images without foreground shadows and real images with foreground shadows. However, due to the training data scarcity and task difficulty, these methods are struggling to generate shadows with reasonable shapes and intensities.

Recently, foundation model (\emph{e.g.}, stable diffusion \cite{Rombach_2022_CVPR}) pretrained on large-scale dataset has demonstrated unprecedented potential for image generation and editing. In previous works \cite{paintbyexample,zhang2023controlcom} on object-guided inpainting or composition, they show that the generated foregrounds are accompanied by  shadows even without considering the shadow issue, probably because of the rich prior knowledge of natural shadow images in foundation model. However, they could only generate satisfactory shadows in simple cases and the object appearance could be altered unexpectedly. 

We build our method upon conditional foundation model \cite{zhang2023adding} and propose several key innovations. First, we modify the control encoder input and the noise loss to fit our task. Then, we observe that the generated shadow intensity (the level of darkness) is unsatisfactory. Especially when the background objects has shadows, the intensity inconsistency between foreground shadow and background shadows make the whole image unrealistic. Therefore, we introduce another intensity encoder to modulate the foreground shadow intensity. Specifically, the denoising U-Net is modified to output both noise map and foreground shadow mask. The intensity encoder takes in the composite image and background shadow mask, producing the scale/bias to modulate the predicted noise within the foreground shadow region. Finally, we devise a post-processing network to rectify the color shift and background variation.

The model training requires abundant pairs of composite images without foreground shadows and real images with foreground shadows. 
The existing real-world shadow generation dataset DESOBA \cite{hong2021shadow} is limited by scale (\emph{i.e.}, 1,012 real images and 3,623 pairs) due to the high cost of manual shadow removal, which is insufficient to train our model. To ensure sufficient supervision, we design a novel data construction pipeline, which extends DESOBA to DESOBAv2 (\emph{i.e.}, 21,575 real images and 28,573 pairs) using object-shadow detection and inpainting techniques. Specifically, we first collect a large number of real-world images with one or more object-shadow pairs. Then, we use pretrained object-shadow detection model \cite{detect4} to predict object and shadow masks for object-shadow pairs. Next, we apply pretrained inpainting model \cite{Rombach_2022_CVPR} to inpaint the detected shadow regions to get deshadowed images. Finally, based on real images and deshadowed images, we construct pairs of synthetic composite images and ground-truth target images.

We conduct experiments on both DESOBAv2 and DESOBA datasets. The results reveal remarkable improvement in shadow generation task, after leveraging the benefits of large-scale data and foundation model. Our main contributions can be summarized as follows: 1) We contribute DESOBAv2, a large-scale real-world shadow generation dataset, which could greatly facilitate the shadow generation task. 2) We propose a cutting-edge diffusion model specifically designed to produce shadows for the composite foregrounds. 3) Through comprehensive experiments, we validate the efficacy of our dataset construction pipeline and the superiority of our proposed model.

\section{Related Work}
\label{sec:formatting}

\subsection{Image Composition}
Image composition aims to overlay a foreground object on a background image to yield a composite result~\cite{lin2018st,wu2019gp, zhan2019adaptive, zhan2020towards, liu2020arshadowgan}. 
Previous research works have tackled different issues that can compromise the quality of  composite images. For instance, image blending methods~\cite{perez2003poisson,wu2019gp,zhang2020deep,zhang2021deep} target at combining the foreground and background seamlessly. Image harmonization methods~\cite{tsai2017deep, xiaodong2019improving, DoveNet2020,cong2021bargainnet,cong2022high} aim to rectify the illumination disparity between foreground and background. Nonetheless, the above methods largely overlook the shadow cast by the foreground onto the background.
%Some methods~\cite{chen2019toward,weng2020misc,zhan2019spatial} seek to address the inconsistencies in geometry, color, and boundary in a unified manner. 
Recently, generative image composition methods \cite{objectstitch,paintbyexample,zhang2023controlcom} can insert a foreground object into a bounding box in the background and the inserted object is likely to have shadow effect. However, they could only generate satisfactory shadows in simple cases and the object appearance could be altered unexpectedly.

\subsection{Shadow Generation}

In this paper, the goal of shadow generation task is generating plausible shadow for the composite foreground. Existing methods can be broadly categorized into rendering based methods and non-rendering based methods. The rendering based methods necessitate a comprehensive understanding of factors like illumination, reflectance, material properties, and scene geometry to produce shadows for the inserted objects. However, such detailed knowledge relies on user input \cite{karsch2014automatic, 2014Exposing, liu2017static,sheng2022controllable,sheng2023pixht} or model prediction~\cite{liao2019illumination,gardner2019deep,zhang2019all, arief2012realtime}, which is either labor-intensive or unreliable \cite{zhang2019shadowgan}. For example,  \cite{sheng2022controllable,sheng2023pixht} could produce compelling results with user control. However, in the composite image, the lighting information should be inferred automatically from background instead of requested by users. 

Non-rendering based methods~\cite{zhang2019shadowgan,liu2020arshadowgan,hong2021shadow,meng2023automatic} aim to translate an input composite image without foreground shadow to an output with foreground shadow, bypassing the need for explicit knowledge of the aforementioned factors. For instance, ShadowGAN \cite{zhang2019shadowgan} utilizes both global and local conditional discriminator to enhance the realism of generated shadows. 
ARShadowGAN \cite{liu2020arshadowgan} emphasizes the importance of background shadow and uses it to guide foreground shadow generation. SGRNet \cite{hong2021shadow} encourages the information exchange between foreground and background, and employs a classic illumination model for better shadow effect. The work \cite{meng2023automatic} produces multiple under-exposure images and fuses them to get the final shadow region. DMASNet~\cite{RdSOBA} decomposes shadow mask prediction into box prediction and shape prediction, achieving better cross-domain transferability. 

To the best of our knowledge, we are the first diffusion-based method focusing on shadow generation. 

\begin{figure*}[t]
\centering
\includegraphics[width=1.0\linewidth]{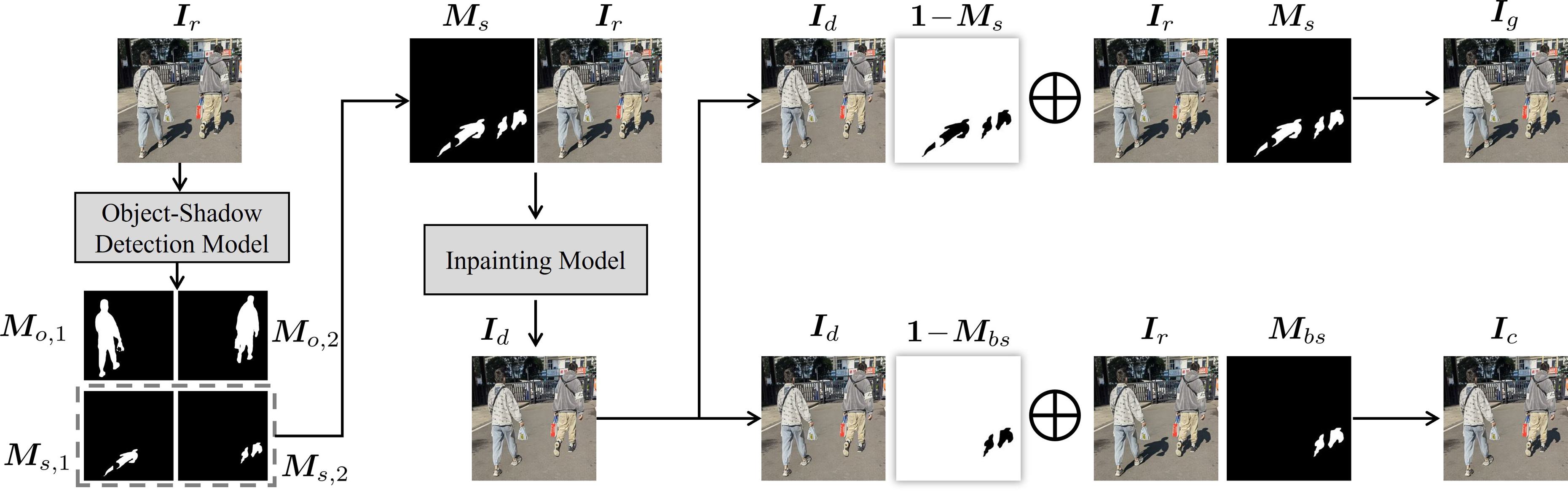}
\caption{The pipeline of dataset construction. We use object-shadow detection model \cite{detect4} to predict pairs of object and shadow masks in the real image $\bm{I}_r$. Then we obtain the union $\bm{M}_s$ of all shadow masks as the inpainting mask and apply inpainting model \cite{Rombach_2022_CVPR} to get a deshadowed image $\bm{I}_d$. After designating a foreground object, we replace the background shadow regions $\bm{M}_{bs}$ in $\bm{I}_d$ with the counterparts in $\bm{I}_r$ to synthesize a composite image $\bm{I}_c$, and replace all the shadow regions  $\bm{M}_{s}$ in $\bm{I}_d$ with the counterparts in $\bm{I}_r$ to obtain the ground-truth target image $\bm{I}_g$.}
\label{fig:dataset_constraction}
\end{figure*}

\subsection{Diffusion Models}
In recent years, diffusion models have emerged as a powerful tool in image generation and image editing. 
These models approach image generation as a series of stochastic transitions, moving from a basic distribution to the desired data distribution~\cite{NEURIPS2020_4c5bcfec}. 
Diffusion models can be divided into unconditional diffusion models~\cite{NEURIPS2020_4c5bcfec, DBLP:journals/corr/abs-2010-02502} and conditional diffusion models~\cite{zhang2023adding,Rombach_2022_CVPR, mou2023t2i}. Unconditional diffusion models focus on generating realistic images by capturing the distribution of natural images, without the need of any specific input conditions. Conditional diffusion models are designed to produce images under the guidance of specific conditional inputs, such as text descriptions, semantic masks, and so on. ControlNet~\cite{zhang2023adding} is a popular conditional diffusion model, which equips large pretrained text-to-image diffusion models with spatial-aware and task-specific conditions. We build our model upon ControlNet and propose several innovations to meet the specific requirements of shadow generation.

\section{Dataset Construction}

The pipeline of our dataset construction is illustrated in Figure~\ref{fig:dataset_constraction}, which will be detailed next.

\subsection{Shadow Image Collection}
We harvest an extensive collection of real-world outdoor images with natural lighting across various scenes from two sources. On one hand, we crawl online images from public websites that have licenses for reuse. On the other hand, we hire photographers to capture photos in the outdoor scenes that satisfy our requirements. We only preserve the images with at least one object-shadow pair, arriving at 44,044 images. 

\subsection{Shadow Removal}
Given a real image $\bm{I}_r$ with object-shadow pairs, we use the pretrained object-shadow detection model \cite{detect4} to predict $K$ pairs of object and shadow masks.  We use $\bm{M}_{o,k}$ (\emph{resp.}, $\bm{M}_{s,k}$) to denote the object (\emph{resp.}, shadow) mask of the $k$-th object.
We refer to one detected object-shadow pair as one detected instance. We eliminate the images without any detected instance. 

Subsequently, we attempt to erase all the detected shadows. We have tried some state-of-the-art shadow removal models  \cite{shadowformer,guo2023shadowdiffusion}, but the performance in the wild is below our expectation due to poor generalization ability. Considering the recent rapid advance of image inpainting \cite{Rombach_2022_CVPR, liu2018partialinpainting, zheng2022image, yang2023uni, huang2023diffusion, parida2023survey} techniques, we resort to image inpainting to remove the shadows. Although image inpainting cannot preserve the background information precisely, we observe that the background textures in the shadow region are usually very simple, and the inpainted result has similar textures with the original background. Thus, we roughly treat the inpainted results as deshadowed results. 

We obtain the union of all detected shadow masks $\bm{M}_{s}=\bm{M}_{s,1}\cup \bm{M}_{s,2} \cup \cdots \cup \bm{M}_{s,K}$ as the inpainting mask and apply the pretrained inpainting model \cite{Rombach_2022_CVPR} to get a deshadowed image $\bm{I}_d$. 
In practice, we observe that the inpainting model is prone to generate low-quality shadow in the inpainted region in some cases. 
To prevent the inpainting model from generating undesirable shadows in the inpainted region, we adopt some tricks like dilating the inpainting mask and flipping images vertically, which can effectively obstruct undesirable shadow generation during inpainting.  However, there may still exist undesirable shadows or noticeable artifacts in the inpainted region. 

After inpainting, we manually filter the object-shadow pairs according to the following rules: 1) We remove the object-shadow pairs with low-quality object masks or shadow masks. 2) We remove those object-shadow pairs with generated shadows or noticeable artifacts in the inpainted region. After manual filtering, we refer to the remaining object-shadow pairs as valid instances. We have 21,575 images with 28,573 valid instances. 

\subsection{Composite Image Synthesis}

\begin{figure*}[t]
\centering
\includegraphics[width=0.85\linewidth]{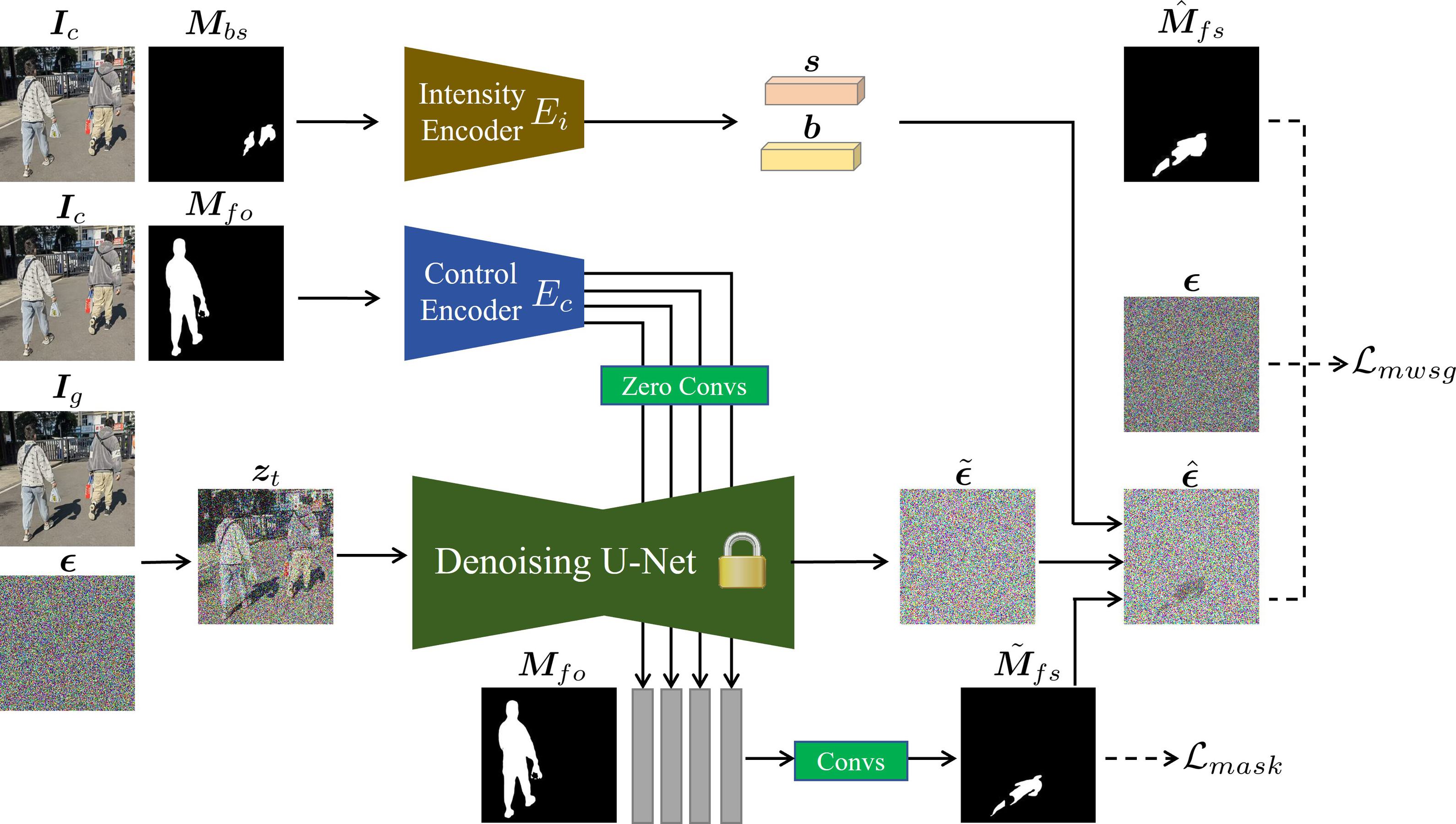}
\caption{The framework of our SGDiffusion. We adapt ControlNet (Control Encoder and Denoising U-Net) to shadow generation task. We also introduce an intensity encoder to modulate the foreground shadow region in the noise map $\tilde{\bm{\epsilon}}$, leading to $\hat{\bm{\epsilon}}$. The output noise $\hat{\bm{\epsilon}}$ is supervised by weighted noise loss $\mathcal{L}_{mwsg}$ based on the expanded foreground shadow mask $\hat{\bm{M}}_{fs}$}
\label{fig:model}
\end{figure*}

Given a pair of a real image $\bm{I}_r$ and a deshadowed image $\bm{I}_d$, we randomly select the $k$-th foreground object from valid instances and synthesize the composite image. $\bm{M}_{o,k}$ (\emph{resp.}, $\bm{M}_{s,k}$) is referred to as the foreground object (\emph{resp.}, shadow) mask $\bm{M}_{fo}$ (\emph{resp.}, $\bm{M}_{fs}$).  One strategy is replacing the shadow region $\bm{M}_{fs}$ of this foreground object in $\bm{I}_r$ with the counterpart in $\bm{I}_d$ to erase the foreground shadow. However, this strategy may leave traces along the shadow boundary, in which case the model may find a shortcut to generate the shadow. 
Another strategy is replacing the shadow regions $\bm{M}_{bs} = \bm{M}_{s,1} \cup \cdots \cup \bm{M}_{s,k-1} \cup \bm{M}_{s,k+1} \cup \cdots \cup \bm{M}_{s,K}$  of the other objects in $\bm{I}_d$
with the counterparts in $\bm{I}_r$ to synthesize a composite image $\bm{I}_c$, in which only the selected foreground object does not have shadow while all the other objects have shadows. We adopt the second strategy. 

After inpainting, the background may undergo slight changes, so the background of $\bm{I}_c$ may be slightly different from that of $\bm{I}_r$. To ensure consistent background, we obtain the ground-truth target image $\bm{I}_g$ by replacing the shadow regions $\bm{M}_{s}$ of all objects in $\bm{I}_d$ with the counterparts in $\bm{I}_r$. Then, $\bm{I}_c$ and $\bm{I}_g$ form a pair of input composite image and ground-truth target image. So far, we obtain tuples in the form of $\{\bm{I}_c, \bm{M}_{fo}, \bm{M}_{fs}, \bm{M}_{bs}, \bm{I}_g\}$, which will be used for model training. Example images and more statistics of our dataset can be found in the supplementary. 
\section{Background}

Stable Diffusion~\cite{Rombach_2022_CVPR} is latent diffusion model operating in a latent space. First, $512\times 512$ images are converted to $64\times 64$ latent images using VAE \cite{kingma2013auto} with encoder $E_r$ and decoder $D_r$. The image space is projected to the latent space using $E_r$, and back to the image space using $D_r$.
Then, the forward diffusion process and backward denoising process are performed in the latent space. The denoising U-Net~\cite{ronneberger2015u} consists of an encoder with $12$ blocks, a middle block, and a skip-connected decoder with $12$ blocks.

During training, random Gaussian noise $\bm{\epsilon}$ is added to the latent image $\bm{z}_0$ in the denoising step $t$, producing a noisy latent image $\bm{z}_t$. Given time step $t$ and text prompt $\bm{c}_{txt}$, the denoising U-Net with model parameters $\epsilon_{\bm{\theta}}$ is trained to predict the added noise $\bm{\epsilon}$.

To support spatial conditional information (\emph{e.g.}, edge, pose, depth), ControlNet \cite{zhang2023adding} integrates a control encoder $E_c$ with pre-trained Stable Diffusion. Specifically, the control encoder contains trainable replicas of its 12 encoding blocks and middle block across four resolutions ($64\times64,32\times32,16\times16,8\times8$).
It takes a $512\times 512$ conditional image as input.

The conditional feature maps $\bm{c}_{img}$ output from control encoder are used to enhance the 12 skip-connections and middle block in denoising U-Net via zero convolution layers. While the original Stable Diffusion is fixed to retain prior knowledge, control encoder could incorporate additional conditions to guide image generation. 

The objective could be rewritten as
\begin{equation}
		\mathcal{L}_{ctrl} = \mathbb{E}_{t,  \bm{\epsilon} \sim \mathcal{N}(0, 1) }\Big[ \Vert \bm{\epsilon} - \epsilon_{\bm{\theta}}(\bm{z}_{t}, t, \bm{c}_{txt}, \bm{c}_{img}) \Vert_{2}^{2}\Big].
		\label{eq:loss_control}
\end{equation}

\section{Method}

Given a composite image $\bm{I}_c$ without foreground shadow as well as the foreground object mask $\bm{M}_{fo}$, our Shadow Generation Diffusion (SGDiffusion) model aims to produce 
$\tilde{\bm{I}}_g$ with plausible foreground shadow. We will adapt ControlNet \cite{zhang2023adding} to shadow generation task in Section~\ref{sec:adapt_controlnet}, and propose novel modules to improve the shadow intensity in Section~\ref{sec:shadow_intensity}. Finally, we will briefly introduce post-processing techniques to enhance the image quality in Section~\ref{sec:postprocessing}.

\subsection{Adapting ControlNet to Shadow Generation} \label{sec:adapt_controlnet}
For shadow generation task, the useful conditional information   is input composite image $\bm{I}_c$ and foreground object mask $\bm{M}_{fo}$, in which the foreground object mask indicates the target object we need to generate shadow for. 
We concatenate $\bm{I}_c$  with $\bm{M}_{fo}$ as the input of control encoder $E_c$. The control encoder outputs the conditional feature maps $\bm{c}_{sg}$, which are injected into the denoising decoder to provide guidance. For the text prompt, we have tried several variants like ``the [object category] with shadow", but they have no significant impact on the generated shadows. Therefore, we use null text prompt by default.

Given a set of conditions including time step $t$ and  conditional feature maps $\bm{c}_{sg}$, the denoising U-Net with model parameters $\epsilon_{\bm{\theta}}$ predicts the noise $\bm{\epsilon}$ added to the noisy latent image $\bm{z}_t$: 
\begin{equation} \label{eqn:L_sg}
	\mathcal{L}_{sg} = \mathbb{E}_{t,  \bm{\epsilon} \sim \mathcal{N}(0, 1) }\Big[ \Vert \bm{\epsilon} - \epsilon_{\bm{\theta}}(\bm{z}_{t}, t,  \bm{c}_{sg})) \Vert_{2}^{2}\Big].
\end{equation}

To enforce the model to place more emphasis on the foreground shadow region, we introduce weighted noise loss, which assigns higher weights to the foreground shadow region. We expand the foreground shadow mask by a dilated kernel to get the expanded mask $\hat{\bm{M}}_{fs}$. The weights in the expanded foreground shadow region are $w$ while the other weights are $1$, leading to the weight map $\bm{W}_{fs}$. If we do not expand the foreground shadow region, the model will be misled to generate large shadows, overlooking the details of shadow shapes and boundaries.
By applying weight map $\bm{W}_{fs}$ to the noise loss, we can arrive at
\begin{equation} \label{eqn:L_wsg}
	\mathcal{L}_{wsg} = \mathbb{E}_{t,  \bm{\epsilon} \sim \mathcal{N}(0, 1) }\Big[ \Vert \bm{W}_{fs}\circ \left(\bm{\epsilon} - \epsilon_{\bm{\theta}}(\bm{z}_{t}, t,  \bm{c}_{sg})\right)\Vert_{2}^{2}\Big],
\end{equation}
where $\circ$ denotes element-wise multiplication.

During inference, to retain more information of input composite image $\bm{I}_c$ in the initial noise, we obtain $\bm{z}_{T}$ by adding noise to the latent image of $\bm{I}_c$, rather than directly sampling from the Gaussian distribution $\mathcal{N}(0, 1)$.

\subsection{Shadow Intensity Modulation} \label{sec:shadow_intensity}
By using the adapted ControlNet in Section \ref{sec:adapt_controlnet}, we observe that the intensity of generated foreground shadow is unsatisfactory. Especially when the background has object-shadow pairs, the generated foreground shadow is often notably darker or brighter than background shadows. Such inconsistency between foreground shadow intensity and background shadow intensity makes the whole image unrealistic.

Therefore, we introduce another intensity encoder to modulate the foreground shadow intensity. Specifically, we use encoder $E_i$ to extract intensity-relevant information. Intuitively, by observing background shadows and its surrounding unshadowed areas, we can estimate the intensity of foreground shadows. Thus, the input of intensity encoder $E_i$ should include the composite image $\bm{I}_c$ and background shadow mask $\bm{M}_{bs}$. When there is no background shadow, the mask is all black. We concatenate $\bm{I}_c$ with background shadow mask $\bm{M}_{bs}$ as the input of intensity encoder.

The intensity encoder outputs scales and biases to adjust the intensity of noise map within the foreground shadow region. The modulated noise map results in the modulated latent image, and further results in the modulated foreground shadow. Therefore, the intensity adjustment of noise map is finally embodied in the intensity variation of generated foreground shadow.   
Specifically, when the noise map has $c$ channels, $E_i$ outputs the $c$-dim scale vector $\bm{s}$ and $c$-dim bias vector $\bm{b}$, containing channel-wise scales and biases. 
$\bm{s}$ and $\bm{b}$ are used to modulate the predicted noise map within the foreground shadow region. 

One problem is that the foreground shadow region is unknown in the testing stage, so we need to predict the foreground shadow mask. To avoid much extra computational cost, we take advantage of the feature maps in the denoising U-Net to predict the foreground shadow mask. 
Previous works usually combine different layers of feature maps in denoising U-Net for mask prediction \cite{xu2023open, ma2023glyphdraw}. We try different layers of feature maps and find that decoder feature maps are more effective in shadow mask prediction. We also use foreground object mask, which could provide useful hints for the location of foreground shadow. We resize all decoder feature maps and foreground object mask to the same size, and concatenate them channel-wisely. The concatenation passes through several convolutional layers to predict the foreground shadow mask $\tilde{\bm{M}}_{fs}$. $\tilde{\bm{M}}_{fs}$ is supervised with ground-truth foreground shadow mask $\bm{M}_{fs}$ by Binary Cross-Entropy (BCE) loss and Dice loss \cite{milletari2016v}:
\begin{equation} \label{eqn:mask_loss}
\mathcal{L}_{mask} =  \mathcal{L}_{bce}(\tilde{\bm{M}}_{fs}, \bm{M}_{fs}) + \mathcal{L}_{dice}(\tilde{\bm{M}}_{fs}, \bm{M}_{fs}).
\end{equation}

When $t$ is large, $\bm{z}_t$ is close to random noise and thus the decoder feature maps are not informative to predict shadow mask. Hence, we only employ the loss $\mathcal{L}_{mask}$ when the time step $t$ is small.  We set the threshold of $t$ as $\sigma T$, in which $T$ is the total number of steps. Accordingly, shadow intensity modulation is only applied when $t$ is smaller than the threshold $\sigma T$.

Provided with the predicted foreground shadow mask $\tilde{\bm{M}}_{fs}$,
we can modulate the noise map within the foreground shadow region.
Given the predicted noise map  $\tilde{\bm{\epsilon}} = \epsilon_{\bm{\theta}}(\bm{z}_{t}, t,  \bm{c}_{sg})$,
we multiply $\tilde{\bm{\epsilon}}$ by channel-wise scales $\bm{s}$ and add channel-wise biases $\bm{b}$ to get $\tilde{\bm{\epsilon}}'$. Then, based on $\tilde{\bm{M}}_{fs}$, we combine the modulated noise map and original noise map to get the final noise map: $\hat{\bm{\epsilon}}=\tilde{\bm{\epsilon}}' \circ \tilde{\bm{M}}_{fs} +  \tilde{\bm{\epsilon}}\circ (1-\tilde{\bm{M}}_{fs})$.

We replace the predicted noise map in Eqn.~(\ref{eqn:L_wsg}) with the final noise map $\hat{\bm{\epsilon}}$ and get
\begin{equation} \label{eqn:L_mwsg}
	\mathcal{L}_{mwsg} = \mathbb{E}_{t,  \bm{\epsilon} \sim \mathcal{N}(0, 1) }\Big[ \Vert \bm{W}_{fs}\circ \left(\bm{\epsilon} - \hat{\bm{\epsilon}}\right) \Vert_{2}^{2}\Big].
\end{equation}

We summarize the mask prediction loss in Eqn.~(\ref{eqn:mask_loss}) and weighted noise loss in Eqn.~(\ref{eqn:L_mwsg}) as
\begin{equation}
	\mathcal{L}_{all} = \mathcal{L}_{mask} + \lambda \mathcal{L}_{mwsg},
\end{equation}
where $\lambda$ is a trade-off parameter.

\subsection{Post-processing} \label{sec:postprocessing}

We observe that the generated images could have color shift and background variation issues. Color shift means that the overall color tone deviates from the input composite image. Background variation means that some background details are changed. To solve these issues, we create a multi-task post-processing network which yields the rectified image together with the foreground shadow mask. Then, we combine input composite image and rectified image based on the predicted foreground shadow mask to produce the final image. The technical details are left to supplementary. 

\section{Experiments}
\begin{figure*}[t]
\centering
\includegraphics[width=1.0\linewidth]{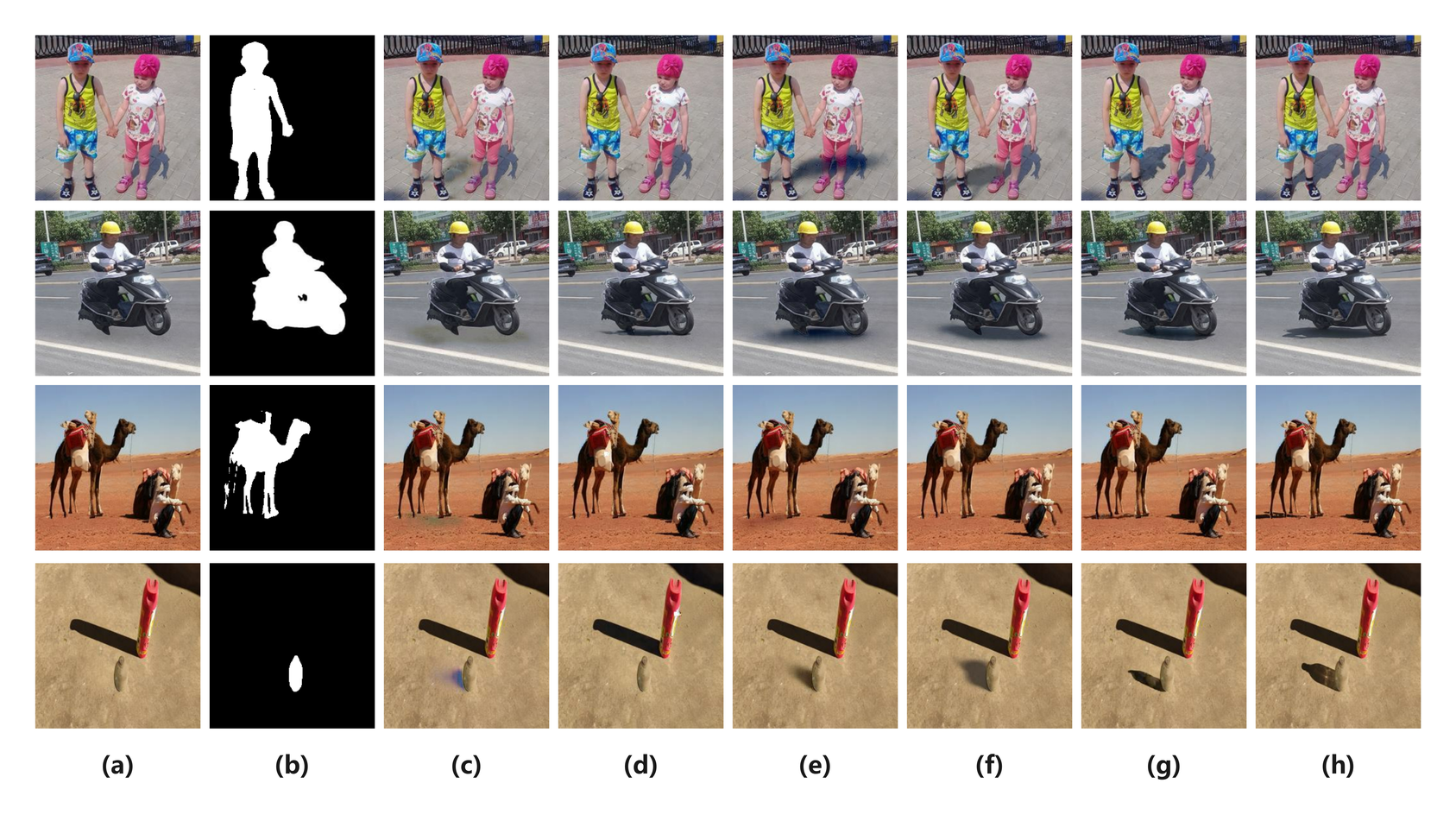}
\caption{Visual comparison of different methods on  DESOBAv2 dataset. From left to right are input composite image (a), foreground object mask (b), results of ShadowGAN \cite{zhang2019shadowgan} (c), MaskshadowGAN \cite{hu2019mask} (d), ARShadowGAN \cite{liu2020arshadowgan} (e), SGRNet  \cite{hong2021shadow} (f), our SGDiffusion (g), ground-truth (h).}
\label{fig:vis_desobav2}
\end{figure*}

\setlength{\tabcolsep}{5pt}
\begin{table*}[t]
  \begin{center}
  %\fontsize{8}{8}\selectfont
    \begin{tabular}{c|cccccc|cccccc}
    %   \Xhline{1.2pt} 
      \toprule[0.8pt]
      \multirow{2}{*}{Method}&
    %   \multirow{2}{*}{Training Set}
      &\multicolumn{5}{c|}{BOS Test Images}
      &\multicolumn{6}{c}{BOS-free Test Images}\cr 
      &GR $\downarrow$ &LR $\downarrow$  &GS $\uparrow$  &LS $\uparrow$ &GB$\downarrow$&LB$\downarrow$&GR $\downarrow$ &LR $\downarrow$  &GS $\uparrow$  &LS $\uparrow$ &GB$\downarrow$&LB$\downarrow$\cr
      \cmidrule(r){1-1} 
      \cmidrule(r){2-7}  
      \cmidrule(r){8-13} 
    ShadowGAN \cite{zhang2019shadowgan} &7.511&	67.464&	0.961&	0.197&0.446	&0.890&17.325& 	76.508& 	0.901 &	0.060 &0.425& 	0.842 \cr 
    MaskshadowGAN \cite{hu2019mask} &8.997&	79.418&	0.951&	0.180&0.500	&1.000&19.338 &	94.327& 	0.906& 	0.044& 0.500& 	1.000 \cr
    ARShadowGAN \cite{liu2020arshadowgan} &7.335&	58.037&	0.961&	0.241&0.383&	0.761&16.067 &	63.713& 	0.908& 	0.104& 0.349 &	0.682\cr 
    SGRNet \cite{hong2021shadow} &7.184&	68.255&	0.964&	0.206&0.301 &0.596 &15.596 &	60.350& 	0.909& 	0.100 &0.271 &	0.534 \cr
    SGDiffusion&\textbf{6.098} 	&\textbf{53.611} &	\textbf{0.971}	& \textbf{0.370} & \textbf{0.245} & 	\textbf{0.487} & \textbf{15.110} &\textbf{55.874} &	\textbf{0.913} &\textbf{0.117} & \textbf{0.233} & \textbf{0.452}\cr
    \bottomrule[0.8pt]
  
    \end{tabular}
    \end{center}
   \caption{The results of different methods on DESOBAv2 dataset. The best results are highlighted in boldface.}
     \label{tab:image_eva}
\end{table*}

\begin{figure*}[t]
\centering
\includegraphics[width=1.0\linewidth]{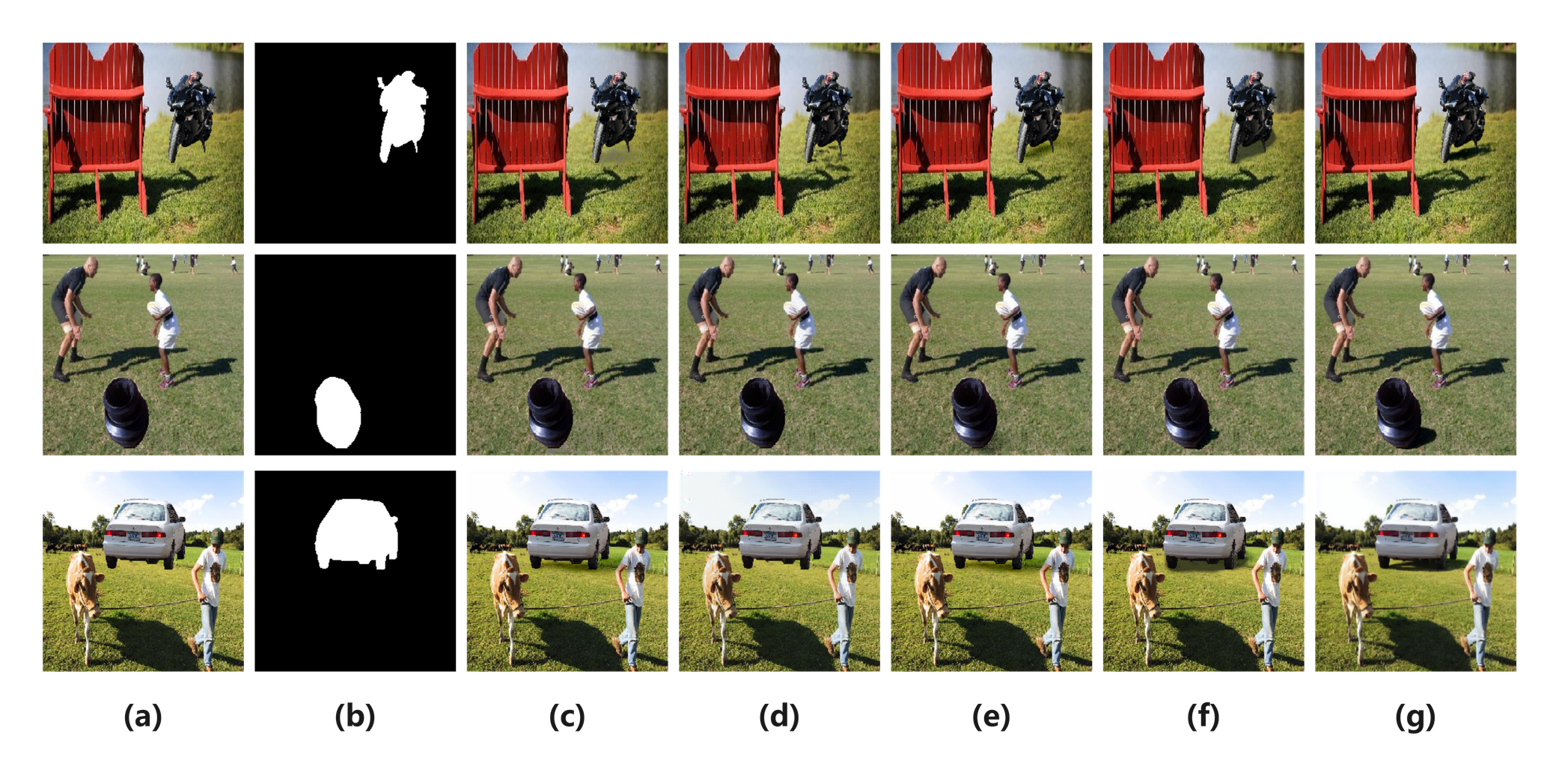}
\caption{Visual comparison of different methods on real composite images. From left to right are input composite image (a), foreground object mask (b), results of ShadowGAN  \cite{zhang2019shadowgan} (c), MaskshadowGAN \cite{hu2019mask} (d), ARShadowGAN \cite{liu2020arshadowgan} (e), SGRNet  \cite{hong2021shadow} (f), SGDiffusion (g).}
\label{fig:vis_real}
\end{figure*}

\begin{table}[t]
  \begin{center}
  %\fontsize{8}{8}\selectfont
  \begin{tabular}{c|ccc|cccccc}
    %   \Xhline{1.2pt} 
      \toprule[0.8pt]
    %   \hline
      Row & WL & IM & PP &GR $\downarrow$ &LR $\downarrow$  &GB$\downarrow$&LB$\downarrow$ \cr
      \hline
    %   \cmidrule(r){2-3}  
    1 & - & - & + & 8.285 & 59.753 & 0.271 & 0.534 \cr
    2 & $\dagger$ & - & + & 8.319 &	59.491 &0.282 &	0.563 \cr
    3 & + & - & + &7.041 &53.829 &  0.249 &	0.492 \cr
    4 & - & $\circ$ & + &7.410 	&56.121&0.269 &	0.536  \cr
    5 & - & + & + & 7.357 &	54.159 &0.262 &	0.526 \cr
    6 & + & + & - &13.447 & 55.231 &0.245& 	0.487\cr
    7 & + & + & + &6.098 	&53.611  &0.245& 	0.487\cr
    \bottomrule
    \end{tabular}
    \end{center}
  \caption{Ablation studies of our method on BOS test images from DESOBAv2 dataset. WL is short for weighted loss and $\dagger$ means without expanding shadow mask. IM is short for intensity modulation and $\circ$ means without using background shadow mask. PP is short for post-processing. }
    \label{tab:ablation}
\end{table}

\subsection{Datasets and Evaluation Metrics}

We conduct experiments on both DESOBA \cite{hong2021shadow} and our contributed DESOBAv2 dataset. We split DESOBAv2 into 21,088 training images with 27,718 tuples and 487 test images with 855 tuples. Following \cite{hong2021shadow}, the test set contains BOS images (with background object-shadow pairs) and BOS-free images. Most of our experiments are based on DESOBAv2 dataset due to the following two concerns: 1) DESOBAv2 has larger test set which supports more comprehensive evaluation. 2) DESOBA has the artifacts caused by manual shadow removal and the existing methods (\emph{e.g.}, SGRNet) tend to overfit such artifacts.  

For the generated results, we evaluate both image quality and mask quality. For image evaluation, following \cite{hong2021shadow}, we adopt RMSE and SSIM, which are calculated based on the ground-truth target image and the generated image. Global RMSE (GR) and Global SSIM (GS) are calculated over the whole image, while Local RMSE (LR) and Local SSIM (LS) are calculated over the ground-truth foreground shadow region. For the mask evaluation, following \cite{hong2021shadow}, we adopt Balanced Error Rate (BER), which is calculated based on the ground-truth binary foreground shadow mask and the predicted foreground shadow mask obtained by threshold 0.5. Global BER (GB) is calculated over the whole image, while Local BER (LB) is calculated over the ground-truth foreground shadow region. Note that diffusion model has stochastic property and 
shadow generation is a multi-modal task, that is, one input has multiple plausible outputs. Similar to multi-modal inpainting evaluation~\cite{zheng2019pluralistic,zhao2020uctgan}, we generate 5 results for one test image with different random seeds and select the one closest to the ground-truth (the highest Local SSIM) to calculate evaluation metrics.

\subsection{Implementation Details}

We develop our method with PyTorch 1.12.1 \cite{paszke2019pytorch}.
Our model is trained using the Adam optimizer \cite{kingma2014adam} with a constant learning rate of $1e^{-5}$ over 50 epochs, on four NVIDIA RTX A6000 GPUs. Our method is built upon ControlNet \cite{zhang2023adding}. We employ ResNet18 \cite{he2016deep} as the intensity encoder. 
The mask predictor passes the concatenation of decoder feature maps and foreground object mask through four convolutional layers, with ReLU activation following the first three layers and Sigmoid activation following the last layer. We set the hyper-parameters $w$, $\sigma$, and $\lambda$ as $10$, $0.7$, and $1$, respectively. %The impact of hyper-parameters will be investigated in the supplementary.  

\subsection{Comparison with Baselines} 

Following~\cite{hong2021shadow}, we compare with ShadowGAN~\cite{zhang2019shadowgan}, Mask-ShadowGAN~\cite{hu2019mask}, ARShadowGAN~\cite{liu2020arshadowgan},  and SGRNet \cite{hong2021shadow}. We train and test all methods on DESOBAv2 dataset. The quantitative results are summarized in Table \ref{tab:image_eva}. We observe that our SGDiffusion achieves the lowest GRMSE, LRMSE and the highest GSSIM, LSSIM, which demonstrates that our method could generate shadow images that are closer to the ground-truth shadow images.
The best GB and LB results demonstrate that the shapes and locations of our generated shadows are more accurate. 

For qualitative comparison, we show several example results in Figure \ref{fig:vis_desobav2}.  Compared with the baseline methods, the shadows produced by our model have more reasonable shapes and intensities. Moreover, as shown in row  $1$, our method can take into account the self-occlusion of objects to generate discontinuous shadows. As shown in row $4$, our method can also consider the material of the objects, producing shadows with translucency effects. We provide more examples in the supplementary.

\subsection{Ablation Studies} 

We study the impact of weighted noise loss (WL), intensity modulation (IM), and  post-processing (PP) of our SGDiffusion on BOS test images from DESOBAv2. The quantitative results are summarized in Table~\ref{tab:ablation}. 

In row 1, we report the results of basic ControlNet without weighted noise loss. 
For WL, the comparison between row 3 and row 1 emphasizes the importance of paying more attention to the foreground shadow region. 
We also report a special case $\dagger$ in row 2, where the foreground shadow mask is not expanded when constructing the weight map. 
The results in row 2 are comparable or even worse than those in row 1, as the model tends to generate larger shadow size while ignoring shape and edge details. 
For IM, the comparison between row 1 and row 5 shows that the intensity modulation can significantly improve the shadow quality by adjusting the shadow intensity.
We also report a special case $\circ$ in row 4, where the intensity encoder input does not contain background shadow mask.
The comparison between row 4 and row 5 shows that background shadow mask is helpful, because the background shadow regions and their surrounding regions could provide useful clues to infer shadow intensity. 
For PP, the comparison between row 6 and row 7 demonstrates that post-processing effectively corrects color shift and background variations, substantially reducing the global RMSE. 
We also provide the visual results of ablated versions in the supplementary.

\subsection{Real Composite Images}
% (main)

We compare different methods on real composite images provided by \cite{hong2021shadow}, where background images and foreground objects are from the DESOBA \cite{hong2021shadow} test set. We train all methods on DESOBAv2 and finetune them on DESOBA. 
The visual results of different methods are showcased in Figure \ref{fig:vis_real}. These results confirm that SGDiffusion adeptly synthesizes lifelike shadows with precise contours, locations, and directions, which are compatible with the background object-shadow pairs and foreground object information.  In contrast, previous methods often produce vague and misdirected shadows. We provide more examples in the supplementary. 

Given the absence of ground-truth images for real composite images, following \cite{hong2021shadow}, we opt for subjective evaluation, engaging $50$ human raters in the user study. Each participant is presented with image pairs from the results generated by $5$ methods, and asked to choose the image with more realistic foreground shadow. Using the Bradley-Terry model \cite{bradley1952rank}, we report the B-T scores in the supplementary, which again proves the advantage of our method.

\section{Conclusion}
In this paper, we have contributed a large-scale shadow generation dataset DESOBAv2. We have also designed a novel diffusion-based shadow generation method.  Extensive experimental results show that our method is able to generate plausible shadows for composite foregrounds, significantly surpassing previous methods. 

\section{Acknowledgement}
The work was supported by the National Natural Science Foundation of China (Grant No. 62076162), the Shanghai Municipal Science and Technology Major/Key Project, China (Grant No. 2021SHZDZX0102). 
{
    \small
    \bibliographystyle{ieeenat_fullname}
    \bibliography{main.bbl}
}


\end{document}

% \maketitle
% \input{sec/7_suppl}
% {
%     \small
%     \bibliographystyle{ieeenat_fullname}
%     \bibliography{main}
% }

\clearpage
\setcounter{page}{1}
\maketitlesupplementary

\begin{figure}[t]
\centering
\includegraphics[width=1.0\linewidth]{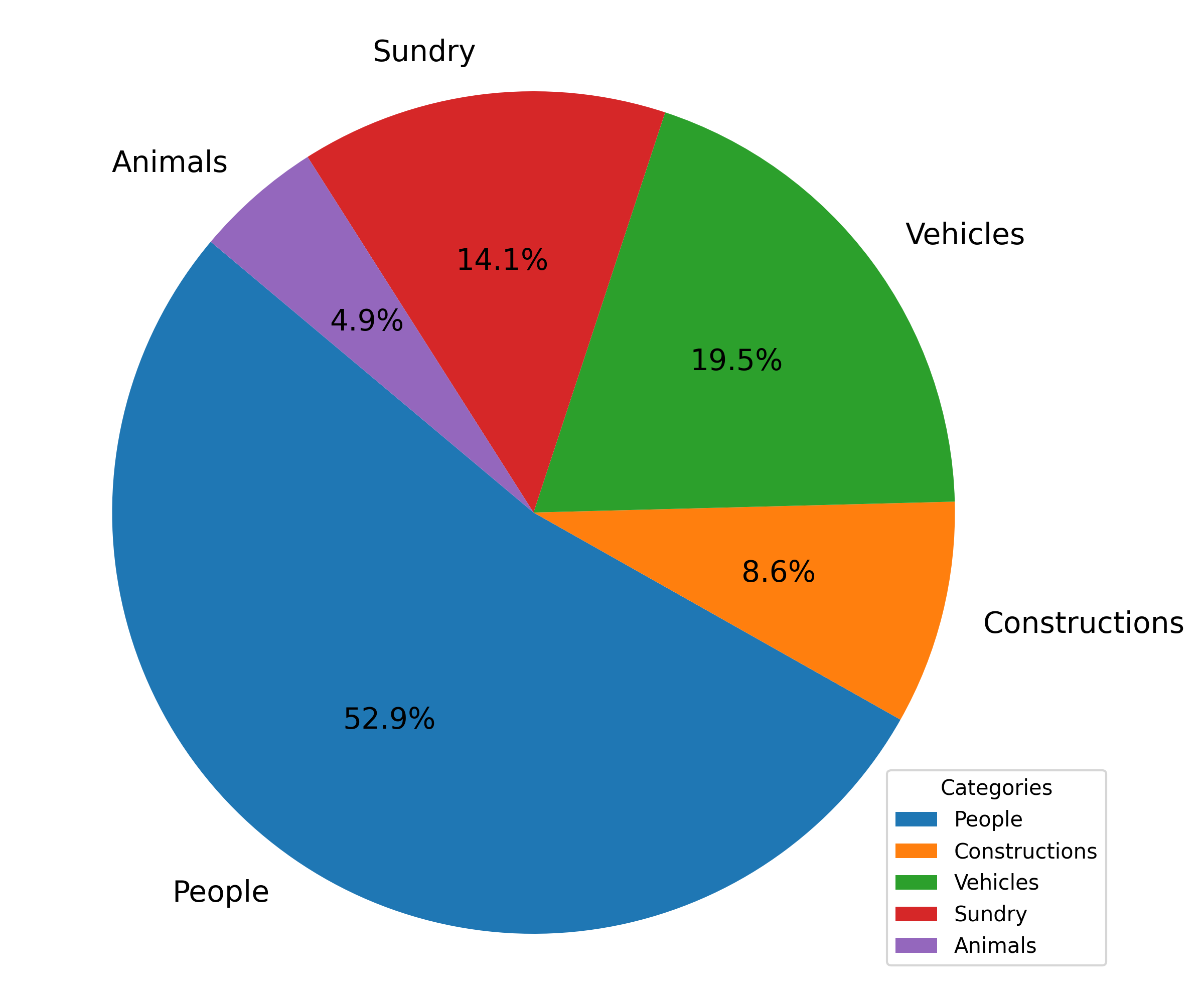}
\caption{The super-category distribution of foreground object in DESOBAv2 dataset.}
\label{fig:category}
\end{figure}

\begin{figure}[t]
\centering
\includegraphics[width=1.0\linewidth]{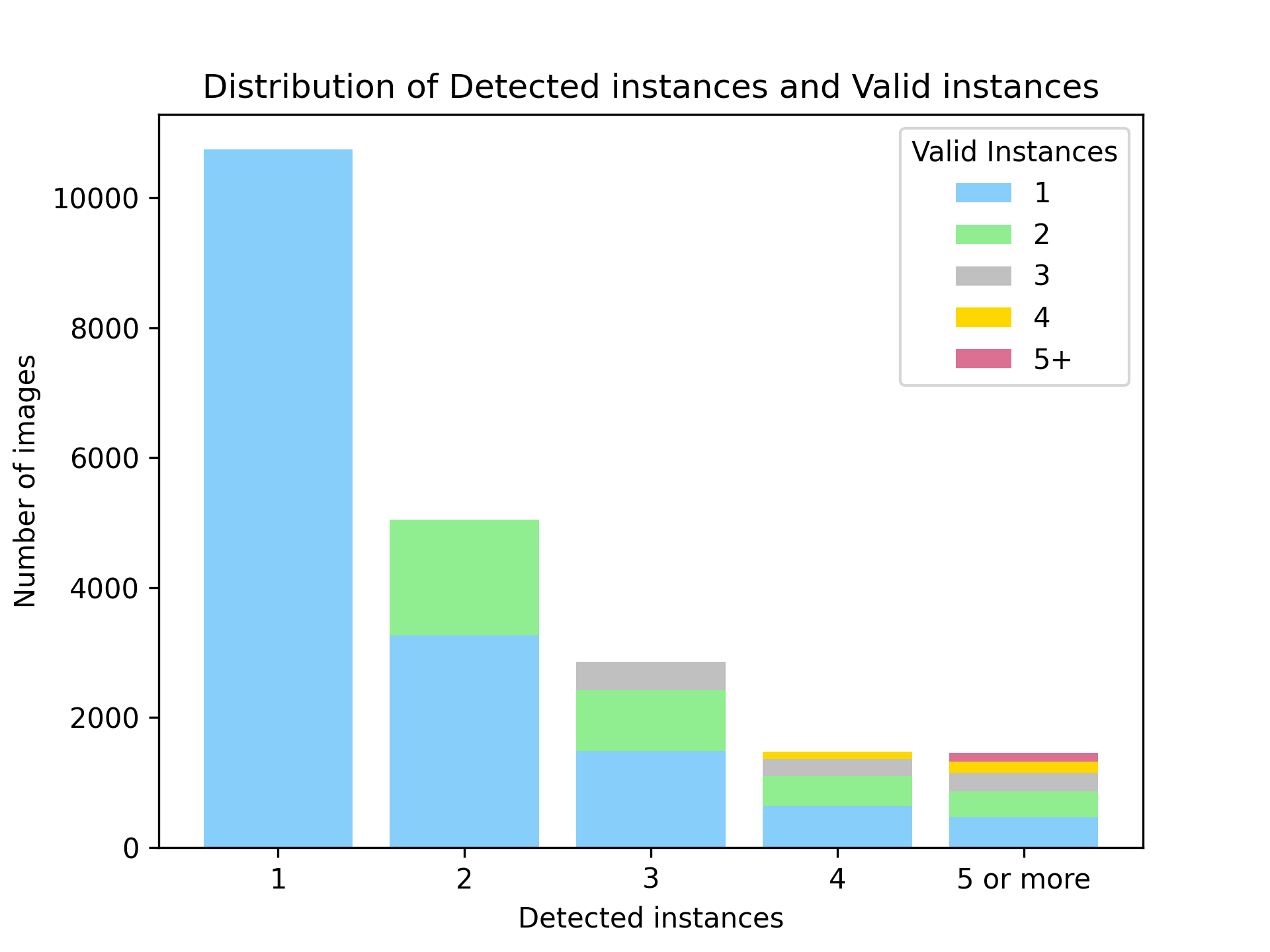}
\caption{The distribution of images with different numbers of detected instances in DESOBAv2 dataset. Each bar further contains the distribution of images with different numbers of valid instances.}
\label{fig:histograms}
\end{figure}

\begin{figure*}[t]
\centering
\includegraphics[width=1.0\linewidth]{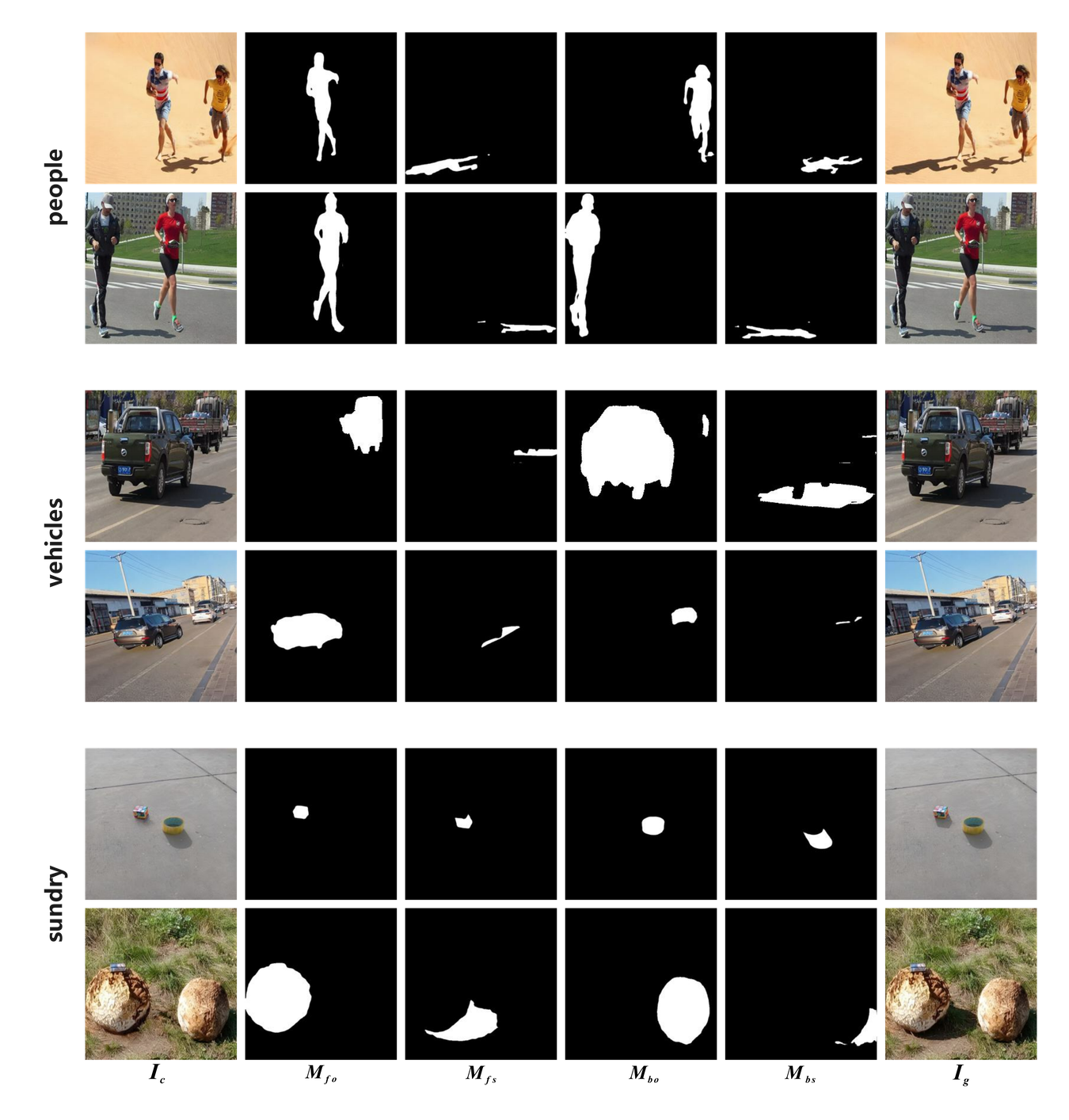}
\caption{Some examples of ``people", ``vehicles", and ``sundry" super-categories in our DESOBAv2 dataset. From left to right in each row, we show the composite image $\bm{I}_c$,  the foreground object mask $\bm{M}_{fo}$, the foreground shadow mask $\bm{M}_{fs}$, the background object mask $\bm{M}_{bo}$, the background shadow mask $\bm{M}_{bs}$, the ground-truth target image $\bm{I}_g$.  }
\label{fig:moreRENOS}
\end{figure*}

\begin{figure*}[t]
\centering
\includegraphics[width=1.0\linewidth]{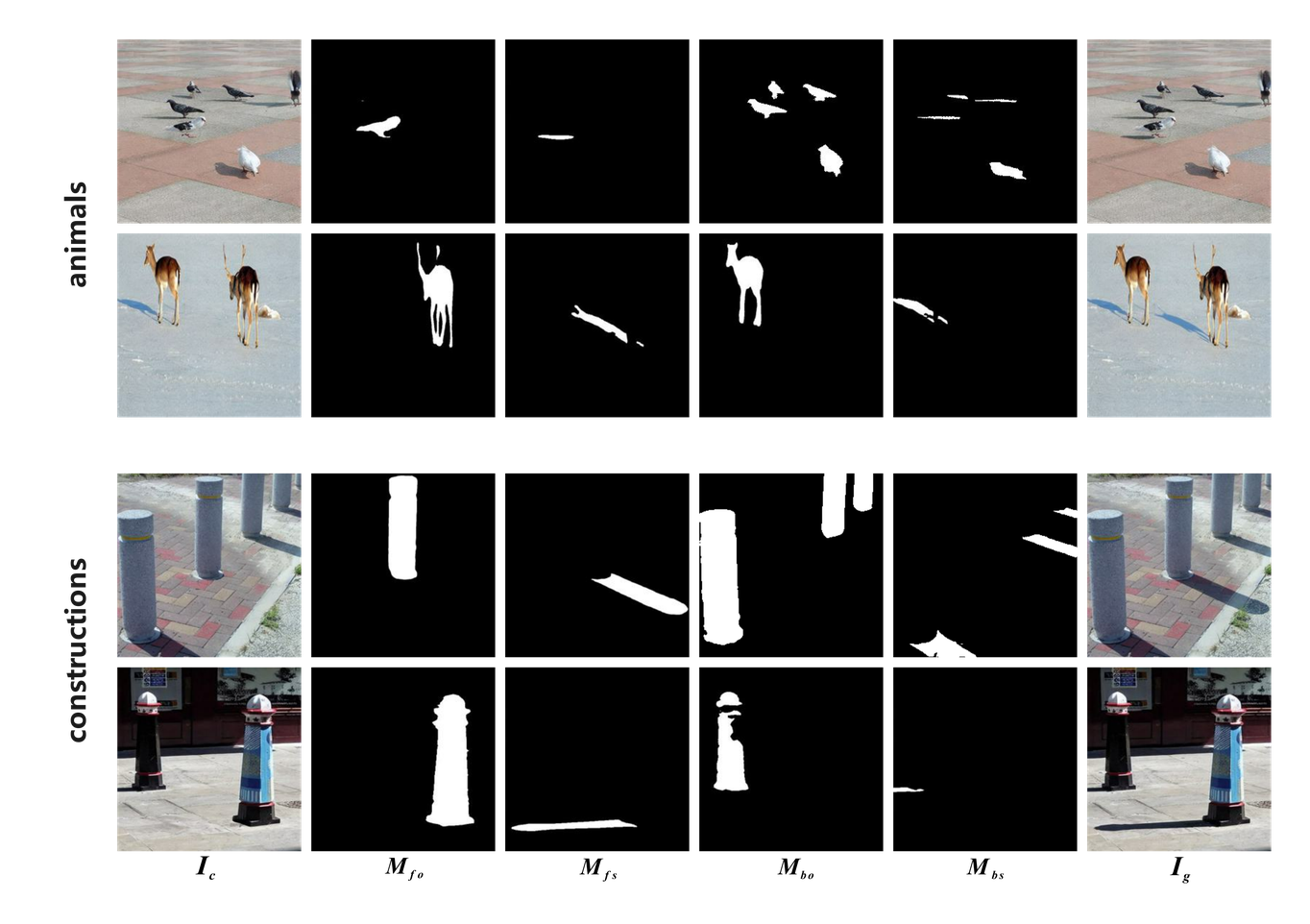}
\caption{Some examples of ``animals" and ``constructions" super-categories in our DESOBAv2 dataset. From left to right in each row, we show the composite image $\bm{I}_c$,  the foreground object mask $\bm{M}_{fo}$, the foreground shadow mask $\bm{M}_{fs}$, the background object mask $\bm{M}_{bo}$, the background shadow mask $\bm{M}_{bs}$, the ground-truth target image $\bm{I}_g$.  }
\label{fig:moreRENOS2}
\end{figure*}

\begin{figure*}[t]
\centering
\includegraphics[width=1.0\linewidth]{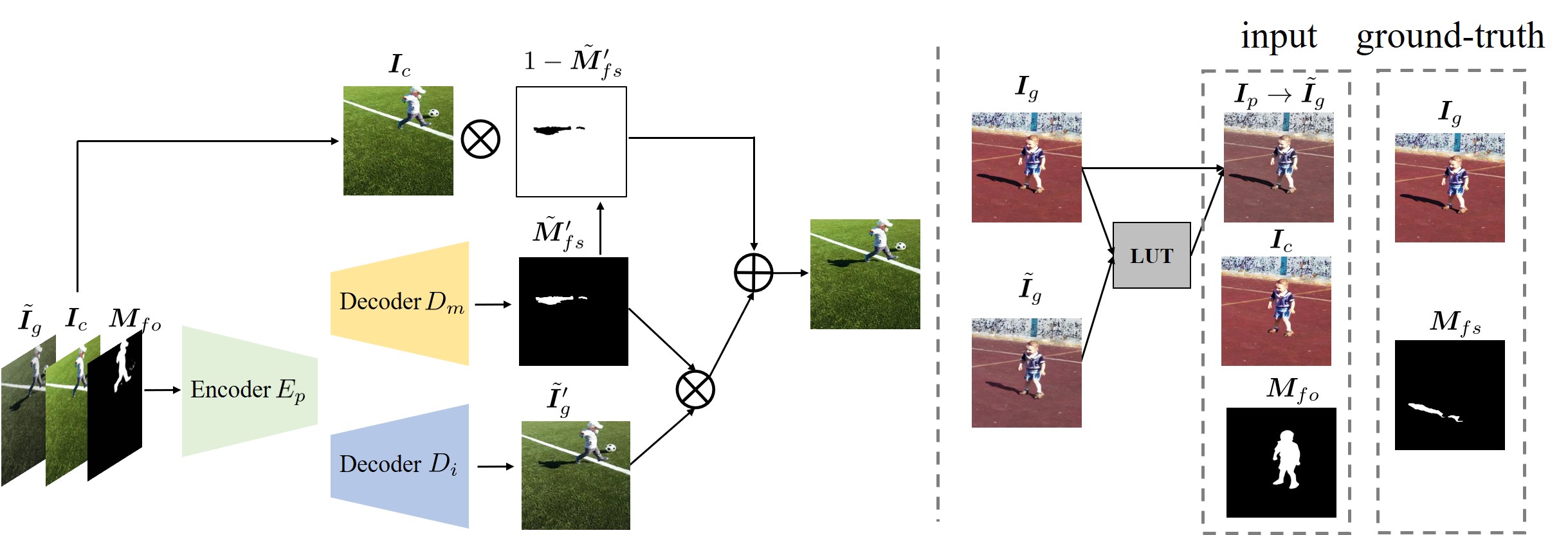}
\caption{The framework of our post-processing network. In the left part, we show our post-processing network structure which can refine a generated image. In the right part, we show the process of constructing training data to train post-processing network. }
\label{fig:post_network}
\end{figure*}  

\begin{figure*}[t]
\centering
\includegraphics[width=1.0\linewidth]{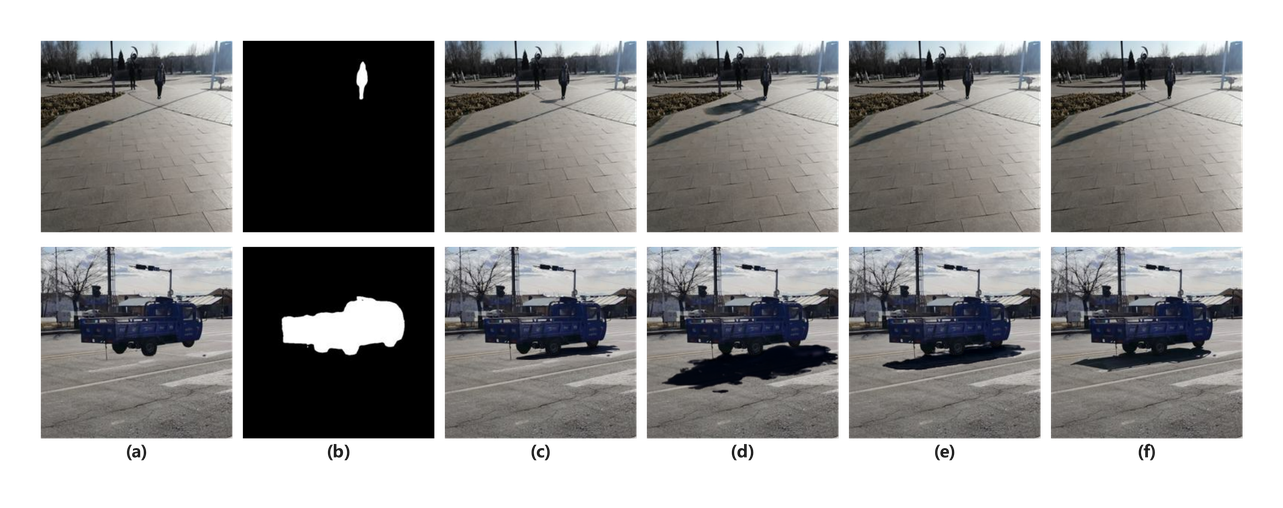}
\caption{The ablation studies on weighted noise loss. From left to right are input composite image (a), foreground object mask (b), results of row 1 (c), row 2 (d), row 3 (e) in Table 2 in the main paper, and ground-truth (f).}
\label{fig:ab_mw}
\end{figure*} 

\begin{figure*}[t]
\centering
\includegraphics[width=1.0\linewidth]{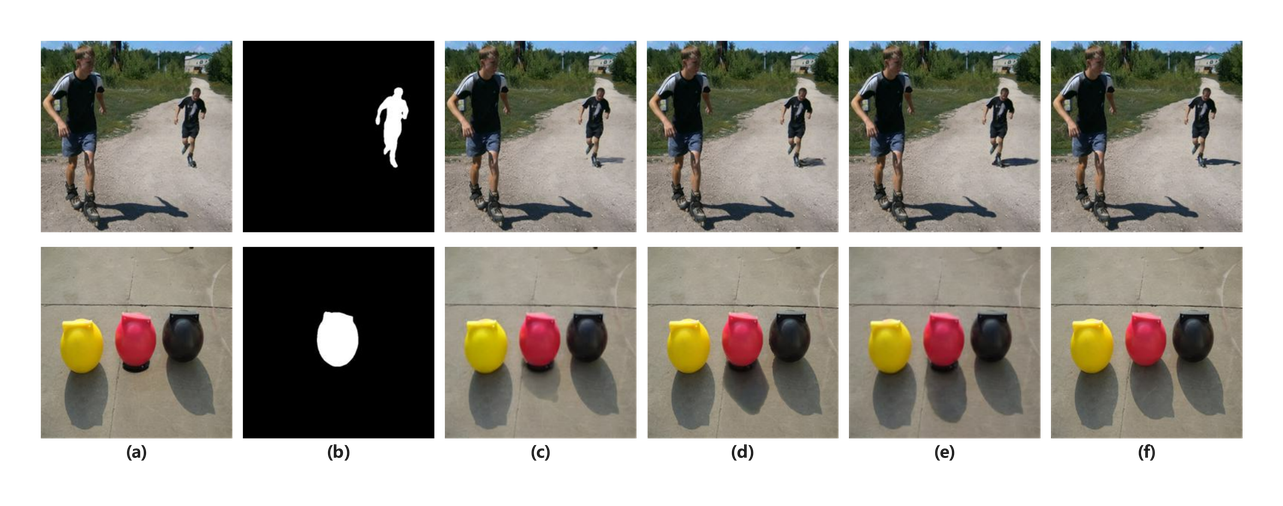}
\caption{The ablation studies on intensity modulation. From left to right are input composite image (a), foreground object mask (b), results of row 1 (c), row 4 (d), row 5 (e) in Table 2 in the main paper, and ground-truth (f).}
\label{fig:ab_im}
\end{figure*} 

\begin{figure*}[t]
\centering
\includegraphics[width=0.9\linewidth]{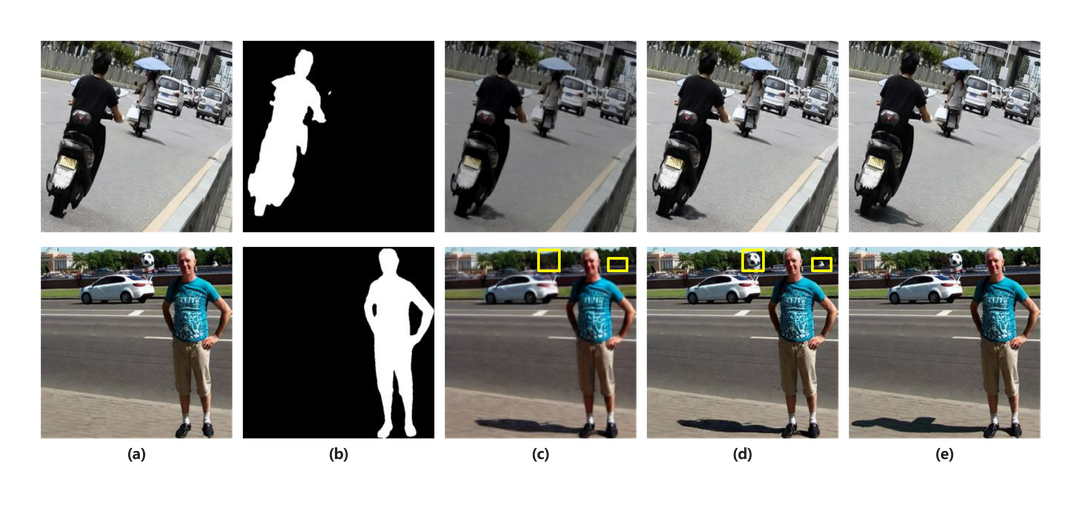}
\caption{The ablation studies on post-processing. From left to right are input composite image (a), foreground object mask (b), results of row 6 (c), row 7 (d) in Table 2 in the main paper, and ground-truth (e).}
\label{fig:ab_pp}
\end{figure*}

In this document, we provide additional materials to support our main submission. In Section~\ref{sec:statistics}, we will provide example images and more statistics of our DESOBAv2 dataset. In Section~\ref{sec:pp}, we will describe the technical details of post-processing. In Section~\ref{sec:vis_ablated_methods}, we will show the results of our ablated versions. In Section~\ref{sec:result_in_v1}, we will show the qualitative results of our method and baseline methods on DESOBA dataset \cite{hong2021shadow}. In Section~\ref{sec:vis_in_v2}, we will show more qualitative results of our method and baseline methods on DESOBAv2 dataset.
In Section~\ref{sec:real}, we will show more qualitative results on the real composite images and report the B-T score. In Section~\ref{sec:Generative}, we will compare our method with recent generative composition methods \cite{objectstitch, paintbyexample}. In Section~\ref{sec:failure_cases}, we will show some failure cases of our method.

\section{More Statistics of Our DESOBAv2 Dataset} \label{sec:statistics}
First, we plot the super-category distribution of foreground objects in our DESOBAv2 dataset in Figure \ref{fig:category}. We classify all objects into five super-categories (people, constructions, vehicles, sundry, animals). From Figure \ref{fig:category}, it can be seen that our DESOBAv2 dataset covers a diversity of categories, in which ``people" is the dominant super-category. 

We provide some examples from our DESOBAv2 dataset in Figure \ref{fig:moreRENOS} and Figure \ref{fig:moreRENOS2}. For each super-category (people, constructions, vehicles, sundry, animals), we show two tuples in the form of $\{\bm{I}_c, \bm{M}_{fo}, \bm{M}_{fs}, \bm{M}_{bo}, \bm{M}_{bs}, \bm{I}_g\}$, in which $\bm{I}_c$ is composite image, $\bm{M}_{fo}$ is foreground object mask, $\bm{M}_{fs}$ is foreground shadow mask, $\bm{M}_{bo}$ is background object mask, $\bm{M}_{bs}$ is background shadow mask, $\bm{I}_g$ is ground-truth target image. 

Then, we summarize the statistics of detected instances and valid instances in our DESOBAv2 dataset. Recall that we use object-shadow detection model \cite{detect4} to detect object-shadow pairs, and refer to one detected object-shadow pair as one detected instance.  After manually filtering the low-quality instances, we refer to the remaining instances as valid instances.
Our DESOBAv2 dataset has in total $21,575$ images. In Figure~\ref{fig:histograms}, we first plot the distribution of images with different numbers of detected instances, based on which most images have fewer than $5$ detected instances. Among the images with specific number of detected instances, we further plot the distribution of images with different numbers of valid instances. 
Note that all images in our dataset have at least one valid instance, so $10,752$ images with one detected instance all have one valid instance. The images with more than one detected instance have different numbers of valid instances.

\section{Technical Details of Post-processing} \label{sec:pp}
To address the problem of color shift and background distortion (see Figure \ref{fig:ab_pp}), we develop a post-processing network which consists of one encoder and two decoders, as illustrated in the left part of Figure~\ref{fig:post_network}. The encoder $E_p$ takes the concatenation of generated image $\tilde{\bm{I}}_g$, composite image $\bm{I}_c$, and foreground object mask $\bm{M}_{fo}$ as input. 
It can be seen that $\tilde{\bm{I}}_g$ and $\bm{I}_c$ have notable color discrepancy. 
One decoder $D_i$ produces the rectified images $\tilde{\bm{I}}'_g$ to fix the color shift problem. Its main goal is adjusting the color of $\tilde{\bm{I}}_g$ to match $\bm{I}_c$, that is, the output $\tilde{\bm{I}}'_g$ and the input $\bm{I}_c$ should be the same except the foreground shadow region. When the color of the other regions is rectified, the color of foreground shadow region is rectified synchronously. The rectified foreground shadow region will be included in the final image. 

The other decoder $D_m$ predicts the foreground shadow mask $\tilde{\bm{M}}'_{fs}$. Intuitively, the foreground shadow region could be easily localized by spotting the content difference between inputs $\tilde{\bm{I}}_g$ and $\bm{I}_c$. The input foreground object mask $\bm{M}_{fo}$ could also provide useful hints for the location of foreground shadow.
Compared with the foreground shadow mask $\tilde{\bm{M}}_{fs}$ predicted by denoising U-Net, $\tilde{\bm{M}}'_{fs}$ is more accurate with higher resolution. The final image can be obtained by $\tilde{\bm{I}}'_g\circ \tilde{\bm{M}}'_{fs} + \bm{I}_c \circ (1-\tilde{\bm{M}}'_{fs})$, in which $\circ$ is element-wise product. In this way, we rectify the color of foreground shadow region and faithfully preserve the background details. 

Both the encoder and the two decoders have four blocks. Each encoder block has three $3\times 3$ conv layers with ReLU followed by a downsampling layer. Each decoder block has three $3\times 3$ conv layers with ReLU followed by an upsampling layer. 
The whole network structure is U-Net, with skip connections from all encoder blocks to the corresponding decoder blocks. 

Next, we discuss the construction process of training data to train the post-processing network. The construction process is illustrated in the right part of Figure \ref{fig:post_network}. We hope that the post-processing network only adjusts the color of $\tilde{\bm{I}}_g$ without changing the foreground shadow shape. 
To simulate the color shift issue, we perturb the color of ground-truth image $\bm{I}_g$ in DESOBAv2 training set. To ensure that the simulated color shift is close to the real color shift, we first obtain our generated result $\tilde{\bm{I}}_g$, and then optimize an image-specific look-up table (LUT) as the color mapping from ground-truth image $\bm{I}_g$ to generated result $\tilde{\bm{I}}_g$. After that, we apply the optimized LUT to $\bm{I}_g$ to get the perturbed ground-truth image $\bm{I}_{p}$. Note that $\bm{I}_g$ and $\bm{I}_{p}$ are only different in color. 
When training the post-processing network, we treat $\bm{I}_{p}$ as the pseudo generated image $\tilde{\bm{I}}_g$, which is taken along with the composite image $\bm{I}_c$ and foreground object mask $\bm{M}_{fo}$ as input. $\bm{I}_g$ is used to supervise the rectified images $\tilde{\bm{I}}'_g$, and  $\bm{M}_{fs}$ is used to supervise the predicted foreground shadow mask $\tilde{\bm{M}}'_{fs}$. 

\setlength{\tabcolsep}{10pt}
\begin{table}[t]
      \begin{center}
  \begin{tabular}{l r}
      % \Xhline{1.2pt} 
      \toprule[0.8pt]
    %   \hline
      Method & B-T score $\uparrow$\cr
    %   \cmidrule(r){2-3}  
    \hline
    ShadowGAN& -0.312 \cr
    Mask-ShadowGAN& -1.284  \cr
    ARShadowGAN& -0.228   \cr
    SGRNet& 0.062 \cr
    SGDiffusion& 1.763 \cr
    \bottomrule[0.8pt]
    \end{tabular}
%   \vspace{0.1mm}
      \end{center}

    \caption{B-T scores of different methods on $100$ real composite images.} 
  \label{tab:bt_score}
\end{table}

\begin{figure*}[h]
\centering
\includegraphics[width=1.0\linewidth]{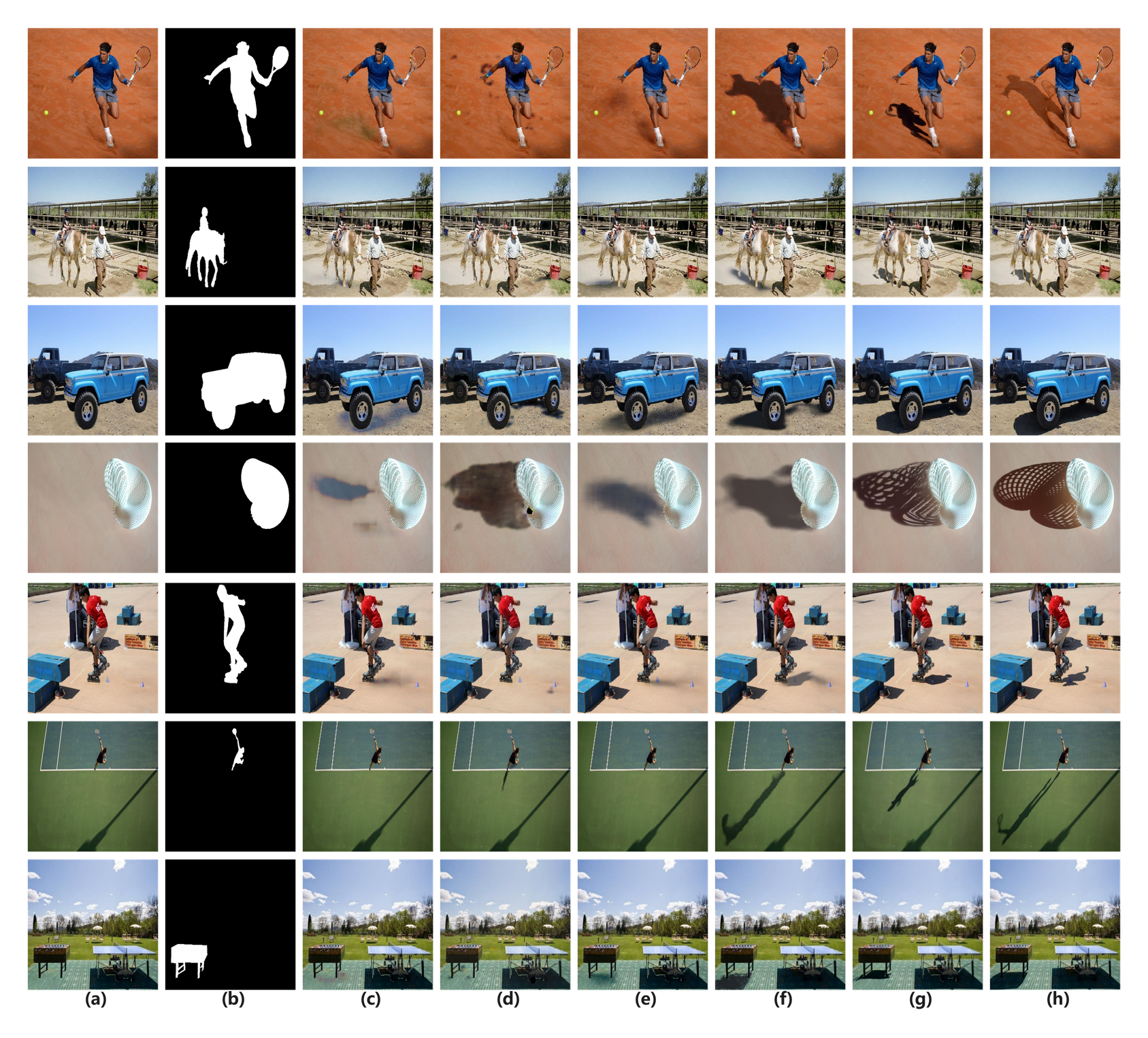}
\caption{Visual comparison of different methods on  DESOBA dataset. From left to right are input composite image (a), foreground object mask (b), results of ShadowGAN \cite{zhang2019shadowgan} (c), MaskshadowGAN \cite{hu2019mask} (d), ARShadowGAN \cite{liu2020arshadowgan} (e), SGRNet  \cite{hong2021shadow} (f), SGDiffusion (g), ground-truth (h).}
\label{fig:v1}
\end{figure*}

\section{Visualization of Ablation Studies}\label{sec:vis_ablated_methods}
In Table 2 in the main paper, we conduct ablation studies to prove the effectiveness of each design in our method. We show the visual results of our ablated versions on DESOBAv2 test set. We divide all ablated versions into three groups. The first group contains row 1, row 2, row 3, which validates the effectiveness of weighted loss. The second group contains row 1, row 4, row 5, which validates the effectiveness of intensity modulation. The third group contains row 6 and row 7, which validates the effectiveness of post-processing. 

The visual results of the first group are shown in Figure~\ref{fig:ab_mw}. By comparing (c) and (e), we can observe that the shapes of generated foreground shadows of (e) are more accurate and closer to the ground-truth (f), which proves that it is useful to pay more attention to the expanded shadow region. However, when not expanding the foreground shadow mask (d), only emphasizing the shadow region makes the model prefer to generate unreasonably large shadows. 

The visual results of the second group are shown in Figure~\ref{fig:ab_im}. By comparing (c) and (e), we can observe that intensity modulation can substantially enhance the intensity of generated shadow, which is more compatible with background shadows. The improvement of (e) over (d) verifies that background shadows could provide useful hints for shadow intensity modulation. 

The visual results of the third group are shown in Figure~\ref{fig:ab_pp}. From (c), we can see that the global color tone of generated image severely deviates from the input composite image (a), and some background details (yellow bounding boxes in the second row) are lost. The improvement of (d) over (c) verifies that post-processing is able to address the color shift and preserve the background details.

\begin{figure*}[t]
\centering
\includegraphics[width=1.0\linewidth]{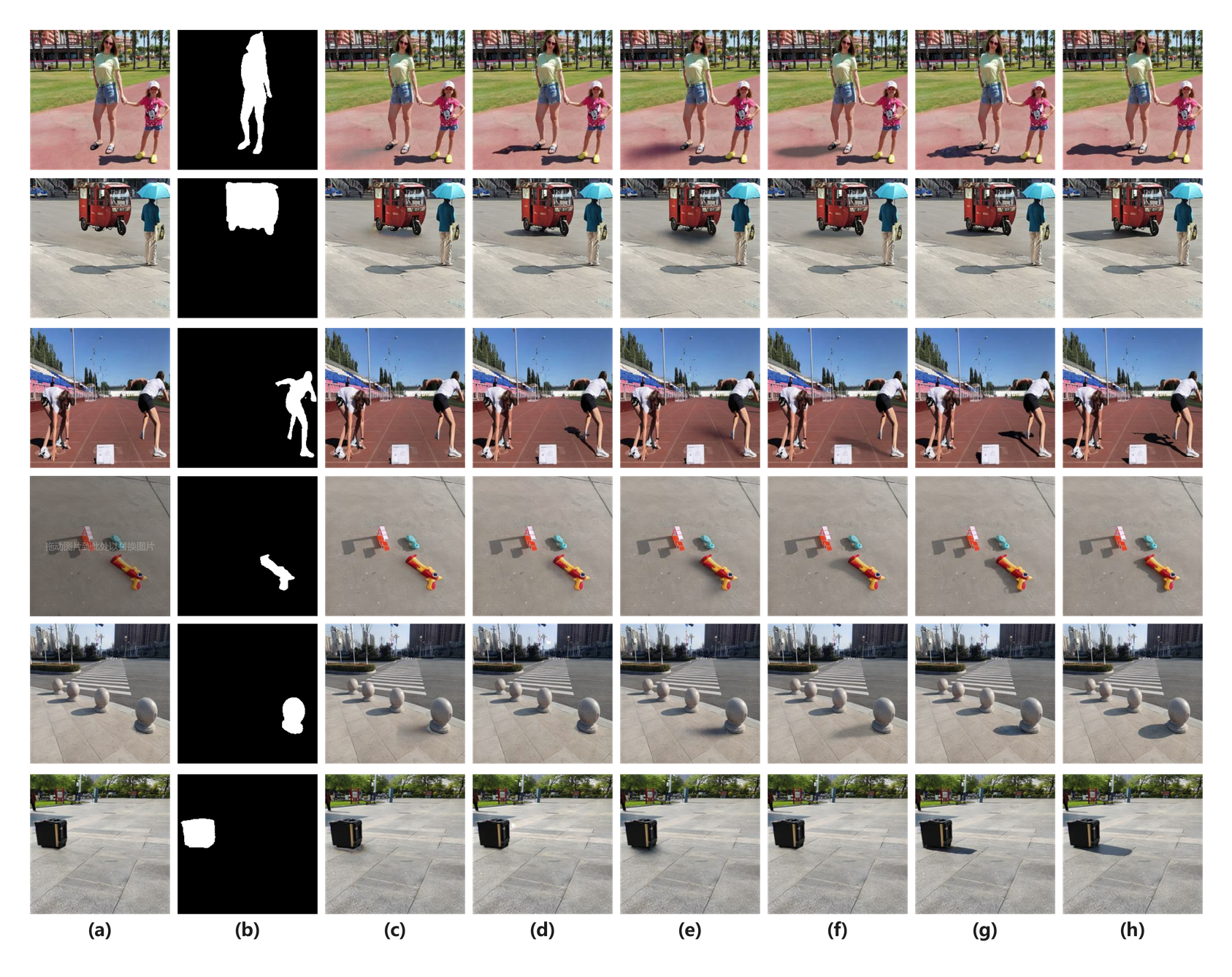}
\caption{Visual comparison of different methods on  DESOBAv2 dataset. From left to right are input composite image (a), foreground object mask (b), results of ShadowGAN \cite{zhang2019shadowgan} (c), MaskshadowGAN \cite{hu2019mask} (d), ARShadowGAN \cite{liu2020arshadowgan} (e), SGRNet  \cite{hong2021shadow} (f), SGDiffusion (g) and ground-truth (h).}
\label{fig:vis_desobav2_supp}
\end{figure*}

\begin{figure*}[t]
\centering
\includegraphics[width=1.0\linewidth]{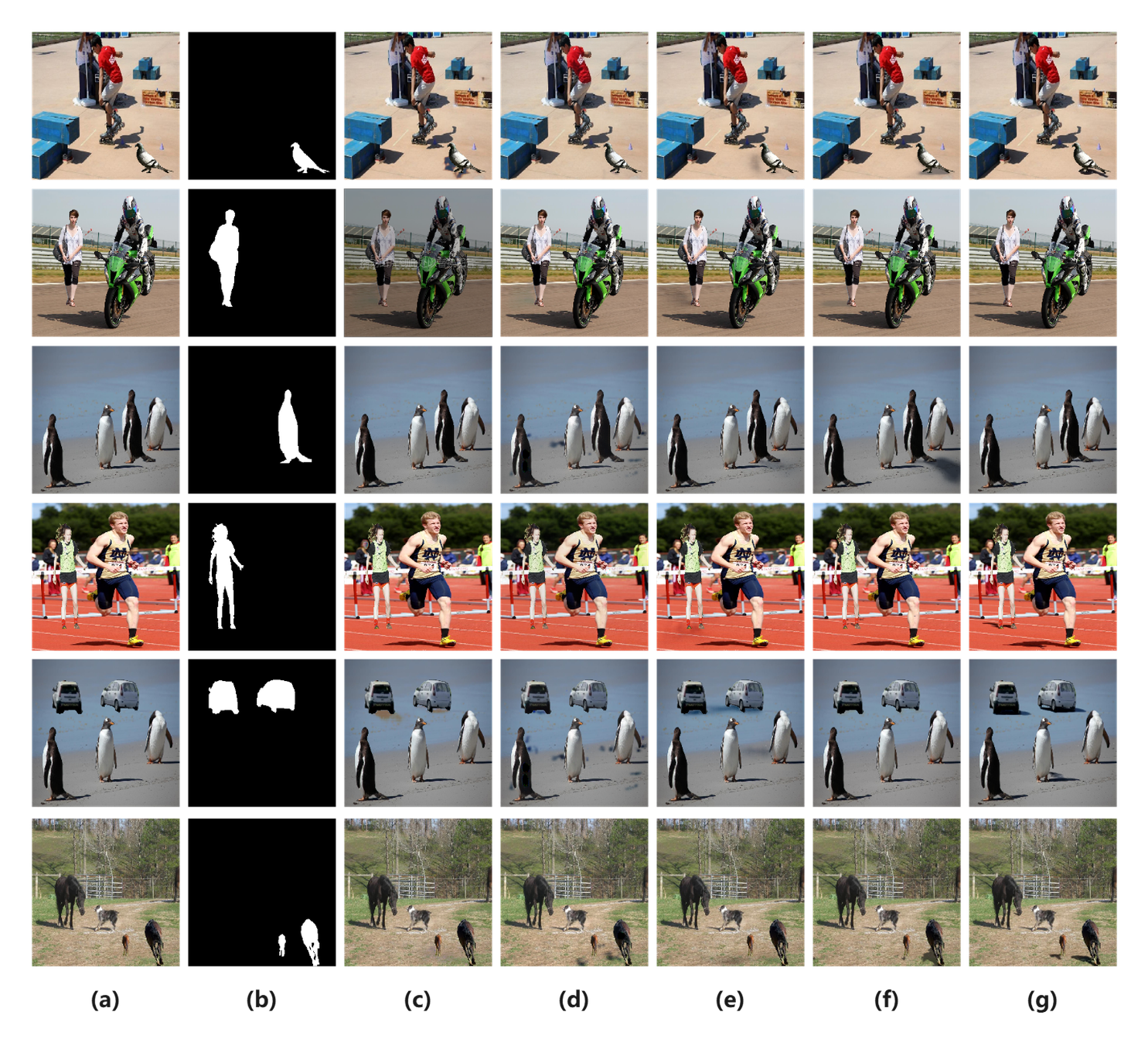}
\caption{Visual comparison  of different methods on real composite images. From left to right are input composite image (a), foreground object mask (b), results of ShadowGAN \cite{zhang2019shadowgan} (c), MaskshadowGAN \cite{hu2019mask} (d), ARShadowGAN \cite{liu2020arshadowgan} (e), SGRNet  \cite{hong2021shadow} (f), SGDiffusion (g).}
\label{fig:vis_real2}
\end{figure*}

\begin{figure*}[t]
\centering
\includegraphics[width=1.0\linewidth]{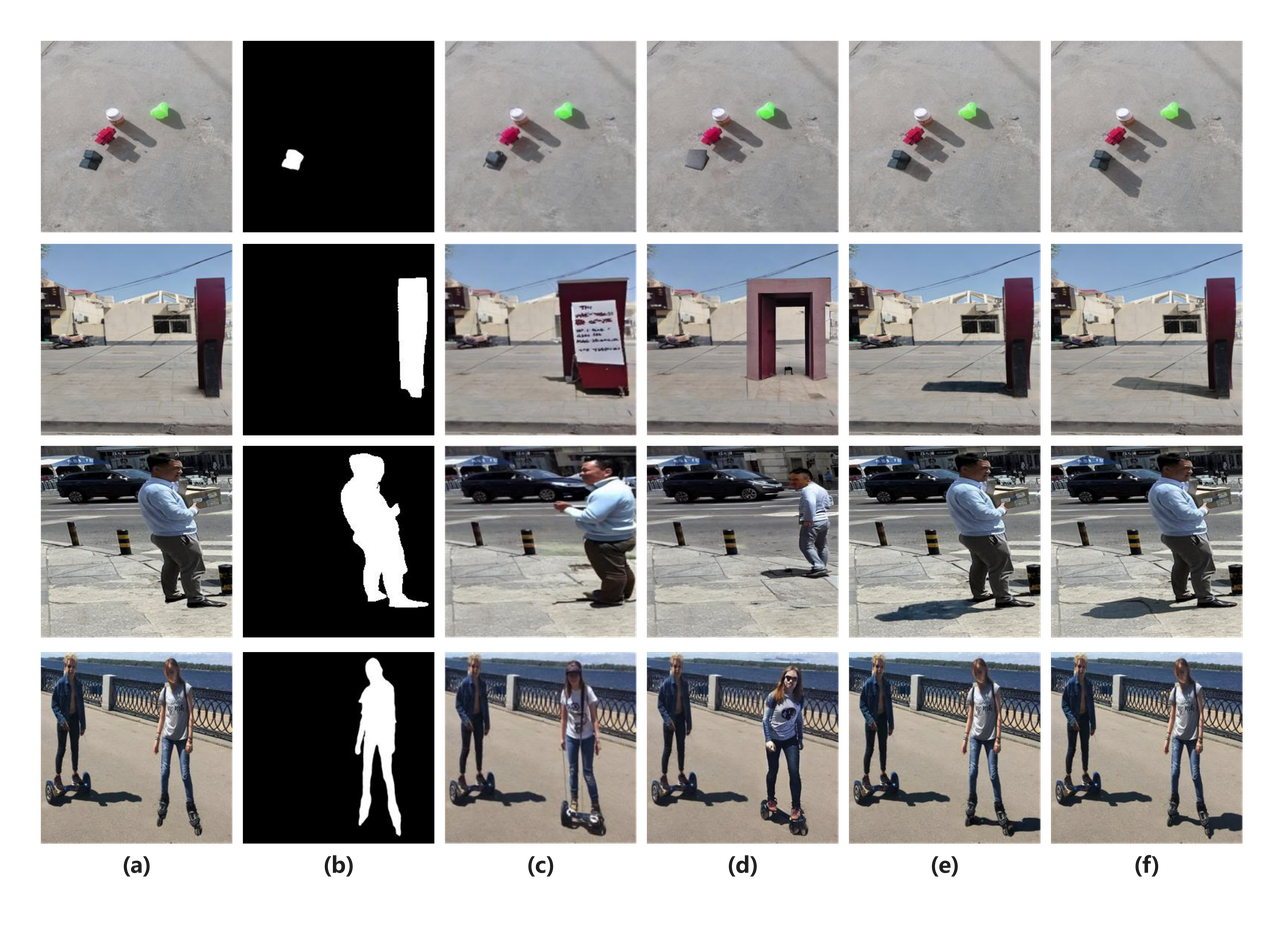}
\caption{Visual comparison with generative composition methods on DESOBAv2 dataset. From left to right are input composite image (a), foreground object mask (b), results of ObjectStitch \cite{objectstitch} (c), PBE \cite{paintbyexample} (d), SGDiffusion (e) and ground-truth (f).}
\label{fig:compare_gen}
\end{figure*}

\begin{figure}[h]
\centering
\includegraphics[width=1.0\linewidth]{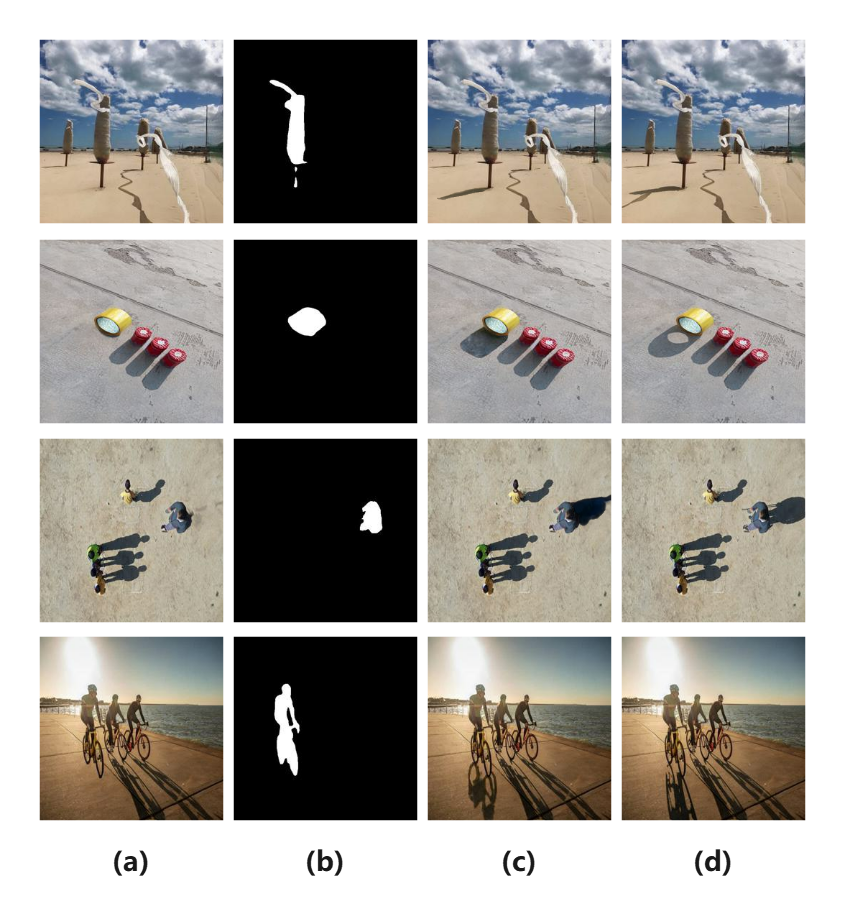}
\caption{Visualization of failure cases produced by our SGDiffusion. From left to right are input composite image (a), foreground object mask (b), results of SGDiffusion (c), ground-truth (d).}
\label{fig:failure}
\end{figure}

\section{Evaluation on DESOBA Dataset} \label{sec:result_in_v1}
To further substantiate the robustness of our model, we train all methods on DESOBAv2 training set and then finetune them on DESOBA training set. We test different methods on DESOBA test set, and visualize the results in Figure~\ref{fig:v1}. 

We can see that our model excels in generating more accurate and plausible shadow shapes in comparison to the baseline methods. 
ShadowGAN \cite{zhang2019shadowgan} and MaskshadowGAN \cite{hu2019mask} are struggling to produce shadows. ARShadowGAN \cite{liu2020arshadowgan} tends to produce oval and blurry shadows, regardless of the object shape.
In contrast, our model is capable of producing shadows with reasonable shapes and intricate details, even for the person with complicated pose (\emph{e.g.}, row $1$) and  hollowed-out objects (\emph{e.g.}, row $4$). 

Besides, we observe that SGRNet is prone to overfit the artifacts in DESOBA dataset. The artifacts are brought by manual shadow removal, based on which the model may find a shortcut for the shadow outline. As shown in row $6$ and row $7$, the generated shadows of SGRNet are surprisingly close to the ground-truth, while the shadow shapes could have many possibilities. We conjecture that SGRNet finds a shortcut based on the artifacts in the foreground shadow regions of input composite images. 

\section{More Visualization Results on DESOBAv2} \label{sec:vis_in_v2}

In the main paper, we have shown the visualization results of different methods on DESOBAv2 test set. 
Here, we provide more visualization results on DESOBAv2 test set in Figure~\ref{fig:vis_desobav2_supp}, based on which we have consistent observations as in the main paper. 

In particular, our method can generate shadows with more plausible shapes and intensity, while other baseline methods even fail to produce any shadow. For example, in row 1 and row 3, the shadow generated by our method has more delicate shape details associated with human pose (\emph{e.g.}, hand holding).

\section{More Results on Real Composite Images} \label{sec:real}

The composite images in DESOBA and DESOBAv2 test sets are synthetic composite images, which may have domain gap with the real composite images. 
To validate the effectiveness of our method on real composite images, we evaluate different methods on $100$ real composite images provided by \cite{hong2021shadow}, which are obtained by pasting foreground objects on background images. Since the foregrounds/backgrounds in real composite images are sourced from DESOBA test set, we train all methods on DESOBAv2 training set and finetune them on DESOBA training set. Note that $100$ real composite images provided by \cite{hong2021shadow} consist of $74$ composite images with one foreground object and $26$ composite images with two foreground objects. For the composite images with two foregrounds, we apply our model twice, each time with the shadow generated for one foreground object.

The results of different methods are shown in Figure~\ref{fig:vis_real2}. Our model is capable of generating realistic foreground shadows, far exceeding the existing baselines. We can observe that the results of SGRNet on real composite images are much worse than those on DESOBA, which again verifies that SGRNet is likely to overfit the artifacts in DESOBA dataset. In contrast, our method has better generalization ability and significantly outperforms SGRNet on real composite images. 

Since these real composite images do not have ground-truth images, we conduct user study to compare different methods. Following SGRNet \cite{hong2021shadow}, given each composite image, we construct 10 image pairs by randomly selecting 2 from 5 results generated by 5 methods. 
In total, we can construct 1000 image pairs based on 100 real composite images.
50 users are asked to participate in this subjective evaluation. Each user could see an image pair each time and select the one whose foreground shadow is more realistic and compatible with the background. In total, 50 users and 1000 image pairs lead to 50,000 pairwise results, based on which the Bradley-Terry (B-T) model \cite{bradley1952rank,lai2016comparative} is used to calculate the ranking of all methods. The B-T scores of different methods are reported in Table \ref{tab:bt_score}. Our method achieves the highest B-T score, which again proves that our method has outstanding generalization ability and achieves better results on real composite images. 

\section{Comparison with Generative Composition Methods} \label{sec:Generative}

With the popularity of generative foundation model, generative image composition has attracted considerable research interest \cite{objectstitch,paintbyexample,zhang2023controlcom}. Specifically, given a pair of background with bounding box and foreground object, they aim to insert the foreground object into the bounding box to produce a realistic composite image, in which the foreground object is seamlessly blended into the background and harmonious with the background. In the generated composite image, the foreground object may have a shadow, even though these methods did not specially consider the shadow problem. 

However, these methods have evident drawbacks. Firstly, they could only insert the object into the specified bounding box, but the shadow could fall out of the scope of bounding box, in which case they are unable to generate reasonable shadow. Secondly, the identity of foreground object could be significantly altered, which may be against the user intention. The comparison between our approach and generative image composition methods \cite{objectstitch,paintbyexample} on DESOBAv2 test set is shown in Figure \ref{fig:compare_gen}. 

For \cite{objectstitch,paintbyexample}, we treat the bounding box enclosing the composite foreground as bounding box and the cropped composite foreground as reference object. We use their released model pretrained on large-scale dataset \cite{openimages}. 
Based on the results in Figure \ref{fig:compare_gen},  we find that these methods \cite{objectstitch,paintbyexample} are not suitable for our task and their generated shadows have poor quality. 
Unlike these methods, our approach can generate shadows with more precise locations and shapes,  while preserving the foreground identity.

\section{Failure Cases} \label{sec:failure_cases}

Our method can generally achieve satisfactory results. However, for some challenging cases, our method may fail to generate plausible shadows. As shown in Figure \ref{fig:failure}, for floating objects, our model often fails to generate their shadows (\emph{e.g.}, the white ribbon in the first row). Additionally, our model sometimes cannot capture the internal structure of certain items (\emph{e.g.}, the hollow center of the rolled-up rug is not reflected in the shadow). Moreover, our model struggles to accurately understand the object shapes from the bird view (\emph{e.g.} row 3). Lastly, when the shadows are long and complex, our model may not produce satisfactory results (\emph{e.g.}, row 4).{
    \small
    \bibliographystyle{ieeenat_fullname}
    \bibliography{supp.bbl}
}